\documentclass[runningheads]{llncs}

\usepackage[T1]{fontenc}
\usepackage{soul}
\usepackage{url}
\usepackage[hidelinks]{hyperref}
\usepackage[utf8]{inputenc}
\usepackage{graphicx}
\usepackage{amsmath}
\usepackage{amssymb}
\usepackage{booktabs}
\usepackage{algorithm}
\usepackage{algorithmic}
\usepackage{caption}
\usepackage{subcaption}
\usepackage[most]{tcolorbox}
\usepackage{fancyvrb}
\usepackage{color,soul}
\usepackage{xcolor}
\usepackage{titletoc}
\makeatletter
\def\verbatim@font{\scriptsize\ttfamily}
\makeatother

\newtcolorbox{promptbox}[1][]{
  colframe=orange!80,
  listing only,
  listing options={
    basicstyle=\ttfamily\small,
    breaklines=true
  },
  breakable,
  #1
}

\newtcolorbox{cbox}[1][]{
  listing only,
  listing options={
    basicstyle=\ttfamily\small,
    breaklines=true
  },
  breakable,
  #1
}

\begin{document}

\title{Ultra Strong Machine Learning: LLM-Generated Explanations Do Not Yet Suffice for Teaching Humans Active Learning Strategy}

\author{Lun Ai\inst{1} \and Johannes Langer\inst{2} \and Ute Schmid\inst{2} \and Stephen Muggleton\inst{1}}
\authorrunning{L. Ai et al.}
\institute{Department of Computing, Imperial College London, UK\\
\email{\{lun.ai15,s.muggleton\}@imperial.ac.uk}
\and
Faculty Information Systems and Applied Computer Science, University of Bamberg, Germany\\
\email{\{johannes.langer,ute.schmid\}@uni-bamberg.de}}

\maketitle

\begin{abstract}
Active learning is a general learning mechanism shared by artificial and human learners. Whether AI can teach humans such a strategy that transfers across domains is an open question. Ultra Strong Machine Learning (USML), a system whose explanations quantifiably improve human out-of-sample performance compared to self-learning, is uniquely positioned to answer this question. Prior USML work relied on hand-crafted explanation templates that require expert effort for each new domain and do not scale. We developed an explanation pipeline combining Inductive Logic Programming (ILP) with large language models (LLMs) to automate explanation generation and scoring. We tested whether these explanations achieve USML in a human trial teaching active learning strategies across three related domains. Our exploratory results show that concise, expert-written explanations benefit learners with higher initial performance, while pipeline-generated explanations provide no advantage over self-learning despite being rated as higher quality from an LLM-as-judge evaluation. This case study reveals a systematic gap that LLM quality metrics do not predict human learning outcomes. Our findings point to explanation complexity relative to task difficulty as a key factor, and call for explanation methods and evaluation criteria grounded in human cognitive constraints rather than LLM preference.
\end{abstract}

\section{Introduction}

\begin{figure}[t]
\centering
\includegraphics[width=0.7\textwidth]{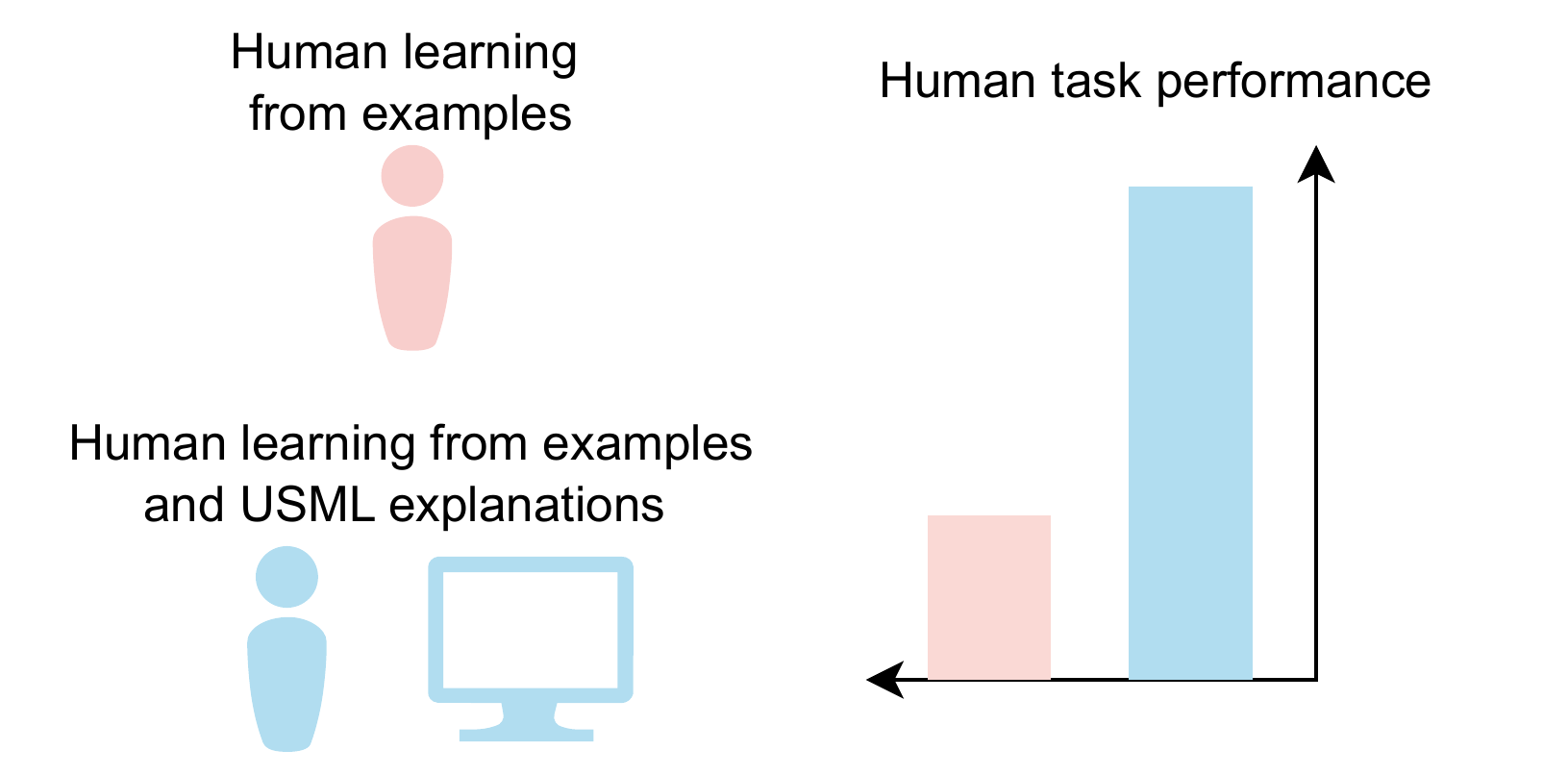}
\caption{USML can quantifiably enhance human task performance compared to human self-learning from examples.}
\label{fig:USML-performance}
\end{figure}

Ultra Strong Machine Learning (USML) \cite{michie1988} distinguishes learning systems by their ability to help humans acquire applicable knowledge. \textit{Weak} machine learning improves performance from training data without an explanation capability. \textit{Strong} machine learning additionally outputs symbolic knowledge for human interpretation. USML (Figure \ref{fig:USML-performance}) builds on strong machine learning, requiring its explanations to quantifiably improve human out-of-sample performance compared to self-learning \cite{Schmid2017}.

Active learning denotes a strategy of selecting the most informative training examples \cite{cohn_improving_1994}, which is a general mechanism shared by artificial and human learners to optimise knowledge acquisition. People naturally seek information that reduces uncertainty \cite{gopnik_scientist_1999,cook_where_2011} and test hypotheses to gain knowledge \cite{popper_logic_2005}, mirroring the computational principle underlying active learning algorithms. Since such transfer learning is central to both human cognition \cite{Anderson1993} and machine intelligence \cite{wiese_mapping_2008,gentner_computational_2011}, this parallel raises a central question: \textit{can AI teach humans an active learning strategy that generalises across domains?} We investigate this through USML, grounding our study in fault diagnosis of electrical circuits \cite{de_kleer_diagnosing_1987}, a canonical diagnostic task where each test partitions the hypothesis space of fault locations and the optimal test maximises hypothesis elimination.

Past work achieved USML in relational learning \cite{US2018}, game playing \cite{Ai2021}, and algorithm discovery \cite{sequential_teaching} using Inductive Logic Programming (ILP) \cite{muggleton_inverse_1995}. ILP learns logic programs that make reasoning steps explicit \cite{michalski_inductive_theory,Anderson1993}, but prior USML systems translated these programs into natural language via hand-crafted templates \cite{Ai2021,sequential_teaching}. This approach requires expert effort for each new domain and does not scale. Large language models (LLMs) offer a natural remedy due to their language generation capability. Together with LLM-as-judge evaluation \cite{li_generation_2025}, LLMs could enable automated explanation generation and scalable quality assessment.

To test whether this approach achieves USML, we develop an explanation pipeline (Figure \ref{fig:explanation_pipeline}) that combines ILP to learn logic programs with LLMs to generate and score natural language explanations. We conducted a human study\footnote{An ethics approval for the empirical study on human participants was issued by the home university's ethics board.} comparing three conditions: 1) human self-learning, 2) human learning with pipeline-generated explanations, and 3) human learning with concise expert-written explanations. Human participants trained on electrical circuits and were tested on two structurally related domains for knowledge transfer.

Our results show that pipeline-generated explanations are rated higher than both direct LLM prompting and hand-crafted templates, yet provide no advantage over self-learning in the human trial. In contrast, concise, expert-written explanations benefit learners with higher initial performance, while pipeline-generated explanations do not despite being rated as higher quality. These findings reveal a systematic gap between LLM quality metrics and human learning outcomes, and highlight the need for explanation methods and evaluation criteria grounded in human cognitive constraints.

\begin{figure*}[t]
\centering
\includegraphics[width=\textwidth]{USML-pipeline-v2.pdf}
\caption{Our explanation pipeline. ILP learns logic programs from examples; coding LLMs interpret programs and reasoning LLMs summarise their outputs into natural language explanations; LLM judges score candidates, optionally using expert-written references.}
\label{fig:explanation_pipeline}
\end{figure*}

\section{Related Work}

\paragraph{Explainable AI.} A great body of work in Explainable AI \cite{gunning_xaiexplainable_2019,barredo_arrieta_explainable_2020} aims to make learning systems more intelligible to humans by providing explanations. Often, little objective evidence \cite{miller_explanation_2019,atzmueller_explainable_2024} can support the benefits of explanations. It is important to take into account how well users understand, for example, from self-reported evaluation \cite{gunning_xaiexplainable_2019,minh_explainable_2022}, but this might be biased by how well the participants understand the domain \cite{schmid_what_2022}. Recent analyses of advanced AI assistants have highlighted that large language models lack explicit representations of human mental states \cite{gabriel_ethics_2024}. This limitation suggests that explanations generated by or optimised for such systems may not align with human cognitive capacities. In our study, we employ an objective evaluation of human understanding by comparing human predictive performance, which is consistent with the aforementioned limitation.

\paragraph{Ultra Strong Machine Learning.} USML was proposed by Michie \cite{michie1988} to characterise systems that improve both machine and human performance. Schmid et al.\ \cite{Schmid2017} operationalised this by studying how predicate invention in ILP affects human comprehension, laying the groundwork for the $E_{ex}$ evaluation framework. The first empirical demonstration of USML used ILP-learned logic programs as direct explanations in a relational learning domain \cite{US2018}. Subsequent work extended USML to game playing \cite{Ai2021}, where hand-crafted visual and textual explanations derived from ILP programs taught humans a winning strategy for the Island Game, and to algorithm discovery \cite{sequential_teaching}, where sequential template-based explanations supported humans in rediscovering merge sort. A consistent finding across these studies is that explanation benefits depend critically on the balance between explanation complexity and task difficulty \cite{Ai2021}, and explanations become counterproductive when their informational content exceeds the cognitive cost of solving the problem unaided. All prior USML systems relied on expert-crafted, domain-specific templates, motivating our investigation of automated explanation generation via LLMs.

\section{Empirical Framework and Methods}

This section introduces the USML problem setting and describes the explanation pipeline used to generate trial materials for the human study.

\subsection{Problem Setting}
\label{sub:problem}

The USML problem is to decide if a learning system can improve human comprehension by explaining what it has learned. We consider the $E_{ex}$ \cite{Ai2021,sequential_teaching} framework. $E_{ex}$ involves a function $D$, a human population $H$ and a set of input-output pairs $E$ of $D$. $H_1, H_2 \subset H$ are randomly sampled groups of similar and sufficient sizes, where $H_1 \cap H_2 = \emptyset$. The unaided human comprehension $\tau_h(D, H_1, E)$ is the mean performance of participants in $H_1$ who, after a brief study of $E$ and without further sight, produce outputs of $D$ from new random samples of its domain. This contrasts with the machine-explained human comprehension, $\tau_{ex}(D, H_2, M(E))$. This is the mean performance of participants in $H_2$ who, after a brief study of the output $M(E)$ from a machine $M$ and without further sight, produce outputs of $D$ from new random samples of its domain. The effect of learning from machine explanations in population $H$ is $E_{ex}(D, H, M(E))$, defined as the difference between the two measurements of comprehension:
\begin{equation*}
E_{ex}(D, H, M(E)) = \tau_{ex}(D, H_2, M(E)) - \tau_h(D, H_1, E)
\end{equation*}

In the context of $D$, $E_{ex}(D, H, M(E)) > 0$ is a necessary condition to conclude the machine $M$ is USML with appropriate statistical tests. This problem has broad applications, as we do not limit the output format of $M$. For example, past work demonstrated USML when the explanations were logic programs \cite{US2018}, natural texts and images \cite{Ai2021,sequential_teaching}. We do not restrict $E_{ex}$ to predictive accuracy since this allows us to compare continuous task performance scores.

\paragraph{Challenges.} Past work \cite{Ai2021,sequential_teaching} explained logic programs from ILP systems based on hand-crafted explanation templates. While effective for human learning, this approach requires expert effort for each new domain and does not scale to more complex logic programs. While we retain expert evaluation for ethics concerns, automated pipelines could support users in creating and selecting accessible explanations for more involved logic programs.

\subsection{Explanation Generation Pipeline}
\label{sec:pipeline}

To generate explanations for the human trial, we develop an explanation pipeline (Figure \ref{fig:explanation_pipeline}) that leverages ILP to learn Prolog programs and employs LLMs with in-context learning to generate explanations through three stages: \textit{learning}, \textit{explanation}, and \textit{scoring}. A set of ILP learning episodes is used as input, and the pipeline produces a set of natural language explanations. Additional human intervention is only required during scoring after the pipeline has been configured with the input episodes, avoiding the need for expert-written explanation templates required by previous USML work.

In the \textit{learning} stage, the pipeline learns logic programs from ILP learning episodes. The typical ILP learning problem involves a set of examples $E = E^+ \cup E^-$, background knowledge $BK$ and relevant parameters for the ILP system. The objective is to find a logic program $h$ such that $\forall e^+ \in E^+ \,\, h \cup BK \models e^+$ and $\forall e^- \in E^- \,\, h \cup BK \not\models e^-$ (``$\models$'' denotes entailment). The pipeline uses an ILP system to solve this learning problem. The learned programs are combined into a ``library'' which subsequent processes can use.

In the \textit{explanation} stage, we leverage two modalities to produce a consensus explanation. We first use multiple coding LLMs to interpret the programs individually. Then, a reasoning LLM summarises code interpretations into a consensus explanation. This is inspired by the ``debate'' strategy \cite{du_improving_2024,subramaniam_multiagent_2024}, which was shown to improve performance in multiple factual reasoning datasets. Each coding LLM is responsible for translating the program into plain English to describe what each part aims to achieve. The reasoning LLMs output natural language explanations that identify the relevant predicates and how these predicates can be applied. We use reasoning LLMs to perform one round of summarisation on the code interpretations to keep the conversation concise with lower token cost.

In the \textit{scoring} stage, we employ LLMs as judges to evaluate summaries using prompts in \cite{zheng_judging_2023}. The LLM judges optionally use expert-written summaries of the ILP-learned programs as references in our study. While LLM judges do not replace human evaluations, we utilise LLM judges as supporting evidence and to score explanations objectively. We adapted their prompt templates for judging single answers with an expert-written reference summary, and examined scores from multiple judges. We believe this mitigates the potential biases of LLMs to favour certain response ordering (position bias) or responses from certain models (self-enhancement bias) \cite{zheng_judging_2023}.

\subsection{ILP Learning}
\label{sub:ilp}

We refer readers to the textbook \cite{ILP_foundation} on logic programming and ILP for background. 
As an ILP approach, Learning from Failures (LFF) \cite{cropper_learning_2021} has a minimal language bias requirement. We use an LFF ILP system Hopper \cite{purgal_learning_2022} to learn second-order programs, which can take other programs as input. Learning with second-order programs can reduce the learned program size, enhance learning performance \cite{cropper_learning_2020} and integrate meta-level and domain-level knowledge \cite{gabaldon_meta-level_2020}.

\begin{figure*}[t]
\centering
\includegraphics[width=\textwidth]{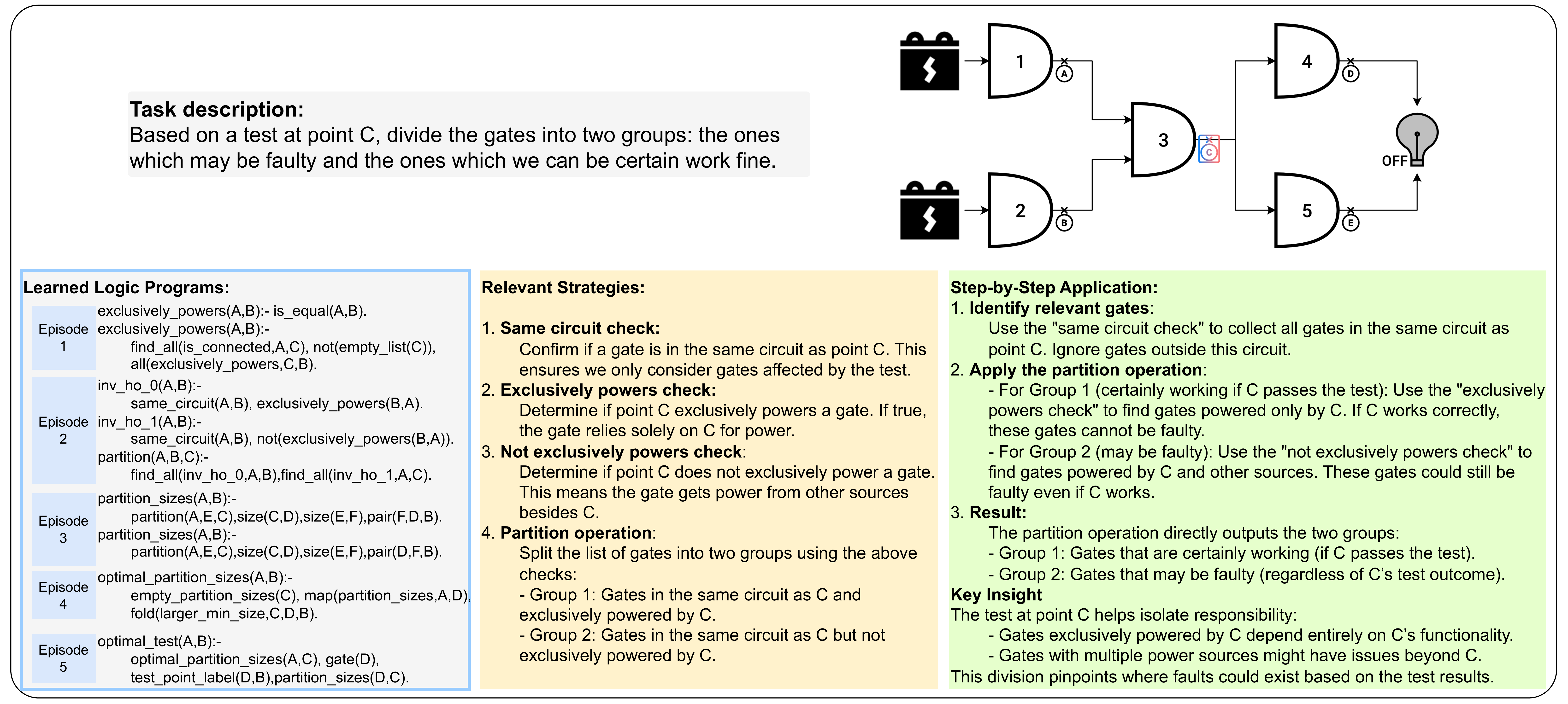}
\caption{The left block shows ILP-learned programs, where each episode is learned from a single circuit example. The middle block summarises relevant programs identified for the task. The right block is an action strategy based on relevant programs.}
\label{fig:explanations}
\end{figure*}

\paragraph{Base Domain.} Our base domain involves identifying the most informative test for diagnosing faults in electrical circuits (Figure \ref{fig:explanations}). All circuits are inspired by an existing network from a biological system (see Appendix \ref{appendix:circuit_design}). Each circuit is powered by batteries and supplies energy to lightbulbs AND gates. If a gate is faulty, it does not transfer energy, and the lightbulbs connected to it would turn off. To find the faulty gate, a power source can be connected to a point of interest, which potentially makes the lightbulb turn on again, therefore giving information about the location of the fault.

\paragraph{Active Learning.} We employ the Hopper system to learn a strategy that resembles a Bayesian active learning approach \cite{hocquette_how_2018}. Under equal priors, the optimal test maximises expected information gain — the binary entropy $H_b$ over the resulting hypothesis partition. Letting $V_s^+$ and $V_s^-$ denote the subsets of the current hypothesis space $V_s$ consistent and inconsistent with the next observation $x_{s+1}$, the minority fraction $p(x_{s+1}, V_s) = \min(|V_s^+|,|V_s^-|)/|V_s|$ gives:
\begin{equation}
\label{eq:information_gain}
\mathbb{E}_{h\sim D_{\mathcal{H}}}[I(x_{s+1}, V_{s}, h)] = H_b(p(x_{s+1}, V_s))
\end{equation}

In our active learning setting, we consider a finite set of potential fault locations, where each location is represented as a single atomic hypothesis. All hypotheses are assigned equal prior probability due to the same textual complexity. A locally optimal test should ideally eliminate approximately half of the hypotheses since it provides the maximum information gain and could reduce the number of subsequent tests needed \cite{mitchell_generalization_1982}. To isolate the effects of hypothesis reduction, we assume that all circuit tests have uniform cost. This cost model supports our USML setting, where we aim to help learners acquire general and transferable strategies.

Hopper learns five sequential episodes \cite{lin_bias_2014}, each building on the previous using a single circuit as training data. The five invented predicates are illustrated in Figure~\ref{fig:explanations}; the complete ILP-learned Prolog programs, episode descriptions, and Hopper inputs are provided in Appendix~\ref{appendix:hopper_input}.

\subsection{LLM Explanation}
\label{sub:llm}
\begin{figure*}[t]
\centering
    \begin{subfigure}{0.45\textwidth}
        \centering
        \resizebox{1.12\textwidth}{!}{\input{figs/circuit/program_scaffolding.pgf}}
        \caption{Program contexts}
        \label{fig:program_scaffolding}
    \end{subfigure}
    \begin{subfigure}{0.45\textwidth}
        \centering
        \resizebox{1.12\textwidth}{!}{\input{figs/circuit/context_scaffolding.pgf}}
        \caption{Program and textual contexts}
        \label{fig:context_scaffolding}
    \end{subfigure}
    \begin{subfigure}{0.45\textwidth}
        \centering
        \resizebox{1.12\textwidth}{!}{\input{figs/circuit/consensus.pgf}}
        \caption{Consensual responses}
        \label{fig:consensus}
    \end{subfigure}
    \caption{Distribution of LLM judged scores for electric circuit domain explanations. RMs and CMs denote reasoning and coding LLMs, respectively. The significance of results has been highlighted by: $p<0.05$ (*), $p<0.01$ (**), $p<0.001$ (***).}
    \label{fig:circuit_score}
\end{figure*}

The explanation stage employs a task-specific approach. Given all ILP-learned logic programs, each reasoning LLM identifies predicates most relevant to solving a provided task. For the human trial, we define three tasks for teaching, each accompanied by a textual description of its objective. The first task involves identifying exclusive sources of power in a circuit. The second task focuses on partitioning the hypothesised fault locations into two groups and computing their sizes. The third task aims to identify the locally optimal test point that yields the most balanced partitions of hypotheses.

Explanations should be understandable to individuals without a technical background. Each reasoning LLM is prompted to generate a clear and concise explanation and invent intuitive names for the relevant predicates. In the scoring stage, explanations are judged based on the above instructions, helpfulness, relevance, accuracy, depth, creativity, and level of detail. We used manually drafted answers as references for judgment (see Appendix~\ref{appendix:human_reference}). The scoring models would justify their scores and penalise overly technical explanations (see Appendix~\ref{appendix:prompt_templates}).

\paragraph{LLM Choices.} We use Qwen2.5 Coder 14B \cite{hui_qwen25-coder_2024} and StarCoder V2 15B \cite{lozhkov_starcoder_2024} as coding LLMs. For judging LLMs, we involve three LLM judges, DeepSeek R1, Claude Sonnet 3.7 and o3-mini, where o3-mini can provide neutral evaluations as it only participates in judging. See Appendix~\ref{appendix:llm_selection} for LLM selection rationale.

\section{Experiments}

We conducted two experiments. The first validates the explanation pipeline by examining whether it produces explanations appropriate for the human trial. The second is the human trial itself, where pipeline-generated explanations are used to teach active learning strategies across three related domains: electric circuits, waterflow and list binary search. Since USML experiments involve humans, relying only on automation or human evaluation could lead to ethical concerns or biases. Therefore, we consider the pipeline as an assistant that generates and judges explanations. Its role is to support human evaluation. We double-check the top-ranking answers to decide the final explanations to be used in the human trial. Before conducting the human trial, we validate that the explanation pipeline produces explanations appropriate for teaching. We address the following questions: \smallskip

\noindent \textbf{Q1.} Do coding models help reasoning models generate higher-rated explanations?

\noindent \textbf{Q2.} Does using multiple coding models improve explanation rating over a single coding model?

\noindent \textbf{Q3.} Does the explanation pipeline provide higher-rated explanations than hand-crafted templates?

\noindent \textbf{Q4.} Do pipeline-generated explanations help humans select more informative examples?

\noindent \textbf{Q5.} Do pipeline-generated explanations help humans perform better at other related tasks?

\subsection{Explanation Pipeline Validation}
\label{exp:pipeline_evaluation}

\paragraph{Generalisability.} We examined pipeline-generated explanations in the base domain and then compared the pipeline against hand-crafted templates (see Appendix~\ref{appendix:templates}) from previous USML work: game playing \cite{Ai2021} and algorithm discovery \cite{sequential_teaching}.

\paragraph{Compared Variables and Conditions.} We evaluate the pipeline across four input configurations: predicate names preserved (NP) or anonymised (AP) to test robustness to naming; a global domain description (GC) or local Prolog circuit examples (LC) as contextual grounding. Baselines omit coding LLMs and directly prompt a reasoning LLM under the same configurations.

\paragraph{Baselines and Methods.} We generated explanations across 14 conditions with three repetitions, and scored each three times per judge, yielding 3024, 2349, and 783 scores for circuits, game playing, and algorithm discovery. We also compared single- versus multi-coding-LLM consensus. Results use one-way ANOVA with post-hoc Tukey's HSD.

\paragraph{Results.} In Figure \ref{fig:program_scaffolding}, LLM judges consider the pipeline with named or anonymised ILP-learned programs superior to direct prompting of a reasoning LLM with anonymised programs. In addition, in Figure \ref{fig:context_scaffolding}, when local or global contexts are present, the pipeline outperforms direct prompting of reasoning LLMs with named or anonymised programs. \textbf{To answer Q1, coding LLMs help reasoning LLMs generate higher-rated explanations}, particularly when programs have less meaningful names.

In Figure~\ref{fig:consensus}, LLM judges consider that using multiple coding LLMs produces better explanations than using just a single coding LLM. \textbf{To answer Q2, both multiple coding models and multiple responses from several models improve explanation rating over a single coding LLM}. This result aligns with past work \cite{du_improving_2024,subramaniam_multiagent_2024} where consensus between multiple LLMs results in higher-quality responses.

Hand-crafted templates from prior work successfully supported USML by translating ILP-learned logic programs into natural language using predefined patterns. While these explanations are more compact and preserve program structure, LLM judges rated them as lacking accessibility and educational value. The insufficient context and depth are reflected in the lower scores of hand-crafted templates compared to pipeline-generated explanations (Appendix Figure~\ref{fig:validation_other}). \textbf{To answer Q3, the explanation pipeline provides higher-rated explanations than hand-crafted templates}.

\paragraph{Human Validation.} We leveraged LLM scoring to filter explanations and guide the final explanation selection by human experts. The top 25\% of explanations is manually evaluated for each task. In the final explanation selection, we value technical correctness over LLM judged scores, as teaching humans incorrect strategies would undermine the learning objectives. We show a selected explanation in Figure~\ref{fig:explanations} with the rest in Appendix~\ref{appendix:selected_explanations} and~\ref{appendix:case_study}. For task 1, we selected the highest-scoring explanation (our explanation pipeline + AP + GC, DeepSeek R1) after confirming no technical errors. For task 2, we selected an explanation (our explanation pipeline + NP, DeepSeek R1) that correctly accounts for the learned strategy. Other high-scoring explanations contain the critical error of ignoring circuit structure. For task 3, we chose an explanation (our explanation pipeline + AP + GC, Claude 3.7 Sonnet) that correctly frames partitioning based on the learned strategy.

\subsection{Human Trial: Does Automated Explanation Achieve USML?}
\label{exp:usml_teaching}

\paragraph{Target Domains}

In the study, we frame the active learning problem described in Section~\ref{sub:ilp} in the following domains. The domain descriptions and visual presentations in the experiment are included in Appendix~\ref{appendix:study_design}.

\textit{Waterflow Domain}. A well-known analogy to the electric circuit base domain (Section~\ref{sub:ilp}) is the waterflow problem \cite{wiese_mapping_2008,gentner_flowing_2014}. The two domains are isomorphic: waterflow circuits and the electric circuits have both component correspondences and functional analogies. The task here is to find exactly one clog in the system by measuring water pressure after circuit nodes.

\textit{List Binary Search Domain}. The task requires finding a number in an ordered list. This domain is isomorphic to the waterflow and circuit domains for ordered lists, since binary search resembles binary partitioning of linear circuits.

\paragraph{Compared Conditions and Design.}
Our study involves a curriculum of three phases (see Appendix~\ref{appendix:study_design}), which correspond to the three tasks described in Section~\ref{sub:llm}.

After the learning phases, participants see a total of 15 trial items equally distributed among the domains. We start with the circuits domain, but randomise the other two to avoid sequence effects. In each item, participants see a visual presentation of the problem according to the domain. Participants must select the first test to perform out of a list of single-choice options, with an ``I don't know'' alternative option. The domains are introduced extensively before the learning phases and each set of trial items.

We map each response to the entropy reduction of a test according to Equation (\ref{eq:information_gain}). Since not every problem contains a test that achieves the maximum entropy reduction of 1, we normalise the maximum achievable score in each trial to 1 while keeping the minimum at 0. This way, a zero score always corresponds to a split with no information gain at all.

Using the symbols introduced in Section~\ref{sub:problem}, we refer to self-learning as the control group ($H_1$) and machine-explained learning as the first treatment group ($H_2$). As a contrast to $H_2$, human-explained learning ($H_r$) is another treatment group, which receives the concise expert-written explanations described in Appendix~\ref{appendix:human_reference} instead of those generated by the pipeline. All groups receive visual feedback on their responses to the items in the learning phases. In addition to their respective explanations, $H_2$ and $H_r$ also see highlights in the visual feedback, which aligns with intermediary program outputs.

\paragraph{Results and Discussion.}
\begin{figure*}[t]
\centering
    \centering
    \resizebox{\textwidth}{!}{\input{fig_exploratory_findings.pgf}}
    \vspace{-25pt}
    \caption{Mean and standard error of human task performance by condition for high-baseline subgroups across domains.}
    \label{fig:exploratory_findings}
\end{figure*}

We recruited 150 participants via the platform Prolific\footnote{\url{https://www.prolific.com}}. We recorded a mean age of 33 with a standard deviation of 10 years, where 79 participants were male and 71 female. We excluded 6 participants due to spending a mean time of more than 120s per trial (population $\mu=62s,\text{std}=32s$). Out of the remaining 144, 40 were in $H_1$, 62 were in $H_2$, and 42 were in $H_r$.

To answer \textbf{Q4} and \textbf{Q5}, we conducted pre-registered analyses testing whether the machine-explained learning treatment affected human task performance across the three domains. We used Kruskal-Wallis H-tests to compare $H_1$, $H_2$, and $H_r$ conditions, with FDR correction applied for each domain. Test results revealed no significant effect of condition on performance in any domain (electric circuits: $H(2)=2.28$, $p=.32$; waterflow: $H(2)=0.33$, $p=.85$; list binary search: $H(2)=1.77$, $p=.41$). \textbf{The answer to both Q4 and Q5 is no. Pipeline-generated explanations of the ILP-learned strategy did not improve participants' ability to select informative examples or transfer learning}.

In addition, we conducted exploratory subgroup analyses to investigate interactions between participants' initial performance and conditions, where initial performance acts as a proxy of participants' ability to adjust to a new domain. In Figure \ref{fig:exploratory_findings}a, we divided participants into high-baseline ($n=56$) and low-baseline ($n=88$) groups using a median split on first-trial performance, with conditions balanced across ability groups ($H=1.92$, $p=.38$). A two-way ANOVA revealed a significant interaction between initial performance and condition in the electric circuits domain ($p=.021$). Among high-baseline participants, Kruskal-Wallis tests indicated a significant condition effect in circuits ($H(2)=8.23$, $p=.016$). Post-hoc Mann-Whitney U-tests with FDR correction showed that expert-written explanations based on the learned programs ($H_r$) significantly outperformed control ($p_{FDR}=.046$, $d=1.08$) and pipeline-generated explanations ($p_{FDR}=.038$, $d=0.97$). Notably, pipeline-generated explanations ($H_2$) did not differ from control ($p=.75$).

We observed effects from further subgroup analyses with reading time (median-split into engaged/quick groups). For list binary search (Figure \ref{fig:exploratory_findings}b), high-baseline engaged readers showed a significant $H_r$ advantage over control ($p_{FDR}=.046$, $d=1.35$) and $H_2$ ($p_{FDR}=.046$, $d=1.31$). For waterflow (Figure \ref{fig:exploratory_findings}c), analysis restricted to difficult trials (mean accuracy $<.70$) revealed a marginally significant advantage in high-baseline quick readers ($H_r$ vs $H_1$: $p_{FDR}=.052$, $d=1.26$; $H_r$ vs $H_2$: $p_{FDR}=.046$, $d=0.93$).

\textbf{While exploratory, these subgroup findings consistently show that higher-rated LLM-generated explanations provide no advantage over self-learning, while concise expert-written explanations benefit learners with higher initial performance.} Expert explanations were brief and focused on the core decision rule, whereas pipeline-generated explanations are more well-structured and verbose. This contrast aligns with cognitive load theory \cite{sweller_cognitive_1988} and prior work \cite{Ai2021} showing that explanations become counterproductive when their informational complexity exceeds the cognitive cost of the underlying task. The LLM judging criteria did not penalise this complexity gap, which accounts for the divergence between high LLM ratings and poor learning outcomes.

\section{Conclusion and Future Work}

Our empirical study reveals key limitations of LLMs as a means to achieve USML, which highlights the need for explanation methods and evaluation metrics grounded in human cognition. This aligns with prior work showing that explanation benefits depend critically on balancing task and explanation complexity \cite{Ai2021}. Future USML research could investigate this balance by modelling human strategy learning through a Bayesian approach to Theory of Mind \cite{tenenbaum_theory-based_2006}, predicting which strategies are likely to be generalised by humans under self-learning and machine-explained conditions.

Our findings also highlight opportunities for improving the explanation pipeline: incorporating recent advances in LLM-based \cite{li_is_2024} and second-order program synthesis \cite{hocquette_learning_2024} could generate explanations for more complex learning tasks that involve meta-level knowledge. The computational framework we establish provides a foundation for such investigations, enabling systematic exploration of when and how automated explanation systems can effectively enhance human learning.

\paragraph{Ethical Statement.} All participants provided informed consent and were compensated for their participation. No personally identifiable information was collected or retained.

\begin{credits}
\subsubsection{\ackname}
The third author acknowledges support from the project ``Learning from Learners'' (VoLL-KI), funded by the German Federal Ministry of Education and Research (BMBF). The authors appreciate suggestions from C\'eline Hocquette and David Cerna on adapting Hopper.
\end{credits}

\bibliographystyle{splncs04}
\bibliography{main}

\appendix
\newpage
\section*{Table of Contents}
\startcontents[appendix]
\printcontents[appendix]{}{1}{\providecommand{\authcount}[1]{}\setcounter{tocdepth}{2}}
\newpage

\section{Pipeline Details}
\label{appendix:pipeline_details}

\subsection{LLM Selections}
\label{appendix:llm_selection}

\paragraph{Coding LLM Choices.} Regarding coding LLM choices, we consider language coverage important for selecting coder models. We use performance in code generation and reasoning as references for LLMs' ability to understand and explain programs. In addition, we selected models of similar sizes since this would not cause a significant difference in model capabilities. Qwen2.5 coder 14B \cite{hui_qwen25-coder_2024} has high performance on the benchmarks for this model class and was trained on 92 programming languages. StarCoder V2 15B \cite{lozhkov_starcoder_2024} was trained on over 600 programming languages and has decent performance on the benchmarks. We used the instruction fine-tuned versions of Qwen2.5 coder 14B and StarCoder V2 15B.

\paragraph{Reasoning LLM Choices.} DeepSeek R1 is a reasoning-focused LLM with open-sourced weights, which has competitive performance in tasks involving reasoning and coding. Claude 3.7 Sonnet is an assistant-focused LLM that has competitive performance on instruction-following, coding, and reasoning benchmarks, with a balance between inference speed, cost, and reasoning power.

\paragraph{Judging LLM Choices.} We employed three LLM judges, DeepSeek R1, Claude 3.7 Sonnet and o3-mini, with each judge evaluating each explanation independently. We use o3-mini as a third-party judge for neutral evaluations. Unlike DeepSeek R1 and Claude 3.7 Sonnet, which both generate and assess answers (self and peer assessments), o3-mini only participates in judging. Judgments from models DeepSeek R1 and Claude 3.7 Sonnet are combined with o3-mini's assessments to mitigate potential self-enhancement bias in LLM self-judgment.

\subsection{LLM Experimental Details}
\label{appendix:llm_experimental}

For interpreting logic programs, we used Ollama with the LangChain Python package to prompt the coding LLMs. The temperature was the default value of 0.8 from the LangChain API. The hardware specification is Google Colab (Ubuntu)/Intel(R) Xeon(R) CPU @ 2.20GHz with 84GB RAM/NVIDIA A100 GPU with 40GB memory. To prompt the reasoning and judging models, we used a temperature of 0.0 to lower the variability of explanation generation and set the reasoning level to medium for o3-mini. The explanations might not be fully reproducible with the same temperatures or future model versions.

\subsection{Usage of Large Language Models}
\label{appendix:llm_usage}
In addition to our empirical study, we used Claude and ChatGPT mainly to polish the language after all intellectual content has been drafted, along with Grammarly as a language editing tool.

\section{Explanation Evaluation On Other USML Tasks}
\label{appendix:generalisation}

\begin{figure*}[h]
\centering
    \begin{subfigure}{0.48\textwidth}
        \centering
        \resizebox{1.13\textwidth}{!}{\input{figs/island/template.pgf}}
        \caption{Game playing \protect\cite{Ai2021}}
        \label{fig:template_island}
    \end{subfigure}
    \hfill
    \begin{subfigure}{0.48\textwidth}
        \centering
        \resizebox{1.13\textwidth}{!}{\input{figs/ms/template.pgf}}
        \caption{Algorithm discovery \protect\cite{sequential_teaching}}
        \label{fig:template_merge_sort}
    \end{subfigure}
    \caption{LLM-judged score distributions comparing pipeline-generated explanations against hand-crafted templates for game playing and algorithm discovery. Annotations and markers are consistent with Figure~\ref{fig:circuit_score} in the main text.}
    \label{fig:validation_other}
\end{figure*}

We applied the explanation pipeline to two additional USML domains: game playing \cite{Ai2021} and algorithm discovery \cite{sequential_teaching}. Across both domains, pipeline-generated explanations consistently receive higher LLM-judged scores than hand-crafted templates from prior work (Figure~\ref{fig:template_island} and Figure~\ref{fig:template_merge_sort}), with statistically significant differences ($p < 0.001$) comparable to those observed for circuit design in the main text. This result indicates that, while hand-crafted templates preserve the syntactic structure of ILP-learned programs, LLM judges consider they lack the contextual depth and accessibility that the LLM explanation pipeline provides. Together with empircal results from the human study, these findings suggest that LLM judges failed to predict the human learning outcomes based on the explanations. This highlights the importance of human evaluation in assessing the educational value of explanations, as LLM judgments alone may not capture the nuances that contribute to effective learning. 

\section{Empirical Study Design Details}
\label{appendix:empirical_study_details}

\subsection{Circuit Design}
\label{appendix:circuit_design}

All circuits were created from an existing metabolic network in biology. A metabolic network is a network of chemical processes that determine the physiological and biochemical properties of a biological system. We used the metabolic network from EcoCyc \cite{moore_revisiting_2024}, which is an online database containing the state-of-the-art biological information about a model organism, \textit{Escherichia coli} (\textit{E.~coli}). We used a collection of key pathways in the metabolic network and their major chemical compounds as connections and gates for building our electric circuits.

\subsection{Study Design Details}
\label{appendix:study_design}

Paragraphs in \textit{italics} are direct quotations from the experimental interface. 


\paragraph{Exemplary Domain Description.}

\textit{
    Your task is to perform tests to detect faults at connections in electrical circuits. Each circuit has one or more sources of energy (battery), lightbulbs (which should always be \textbf{on}), and gates (which are connections between cables). If a gate is faulty, it does not transfer energy, and the lightbulbs connected to it turn off. To find the faulty gate, you can add an additional power source at a test point, which potentially makes the lightbulb turn on again, giving you information about the location of the fault in the circuit. Here is a simple illustration:
}

\begin{figure}[h]
    \centering
    \includegraphics[width=\linewidth]{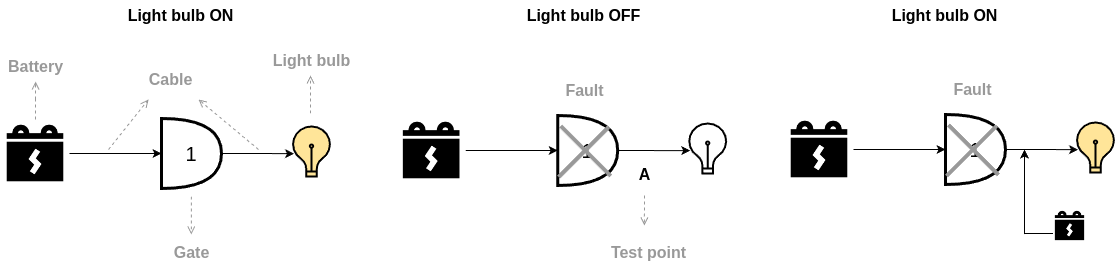}
    \caption{Visual domain introduction used in the study.}
    \label{fig:intro_01}
\end{figure}

\textit{
    These are simplified circuits in that we can always assume that the circuit is closed after the lightbulb and the fault lies "on the path to the lightbulb". Here is a summary of the components of the circuits:
}
\textit{
    \begin{itemize}
        \item Battery: The battery is the "source" of the current going through the circuit.
        \item Cable: The arrows are cables which transfer current. Note that current is only transfered in the direction of the arrows!
        \item Gate: The shapes labeled with numbers (in the example above "1") are gates which connect the current from the incoming arrow to the outgoing arrow. They may be faulty, in that they do not output the current, although they receive current from all input cables.
        \item Lightbulb: The lightbulb turns on if it receives power from all input cables.
        \item Test point: Test points are cables labeled with letters (in the example above "A"). You can perform a test by supplying power to such a cable, effectively ignoring all prior gates.
    \end{itemize}
}
\textit{
    Both the gate and the lightbulb only work if they receive current from all connected input cables (similar to an AND gate). Consider these examples:
}

\begin{figure}[h]
    \centering
    \begin{subfigure}{0.48\linewidth}
        \centering
        \includegraphics[width=0.7\textwidth]{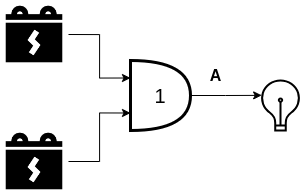}
    \end{subfigure} \hfil
    \begin{subfigure}{0.48\linewidth}
        \centering
        \includegraphics[width=0.7\textwidth]{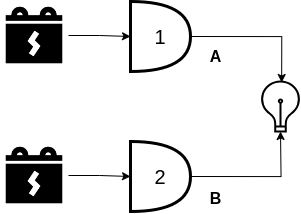}
    \end{subfigure}
    \caption{Demonstration of AND gates used in the study.}
    \label{fig:intro_AND}
\end{figure}

\textit{
    Gate \textbf{1} in the left example is only active if it receives current from both input cables. Similarly, the lightbulb in the right example must receive current from both input cables to turn on.
}

\textit{
    Each circuit you will see in this study has exactly one faulty gate. Your task is to choose the first test to perform so that the fault can be found with the least tests possible, regardless of the result of the test. Here is a worked out example:
}

\begin{figure}[h]
    \centering
    \includegraphics[width=\linewidth]{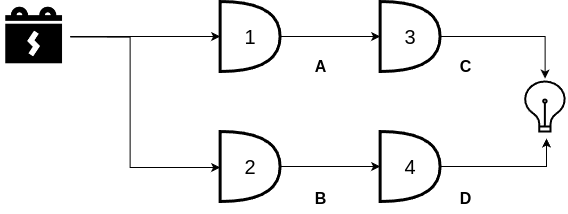}
    \caption{Worked-out example used for the introduction of the circuits domain in the study.}
    \label{fig:intro_wo}
\end{figure}

\textit{
    In this scenario, two tests would be equally viable: \textbf{C} and \textbf{D}. Consider this for test \textbf{C}: If the lightbulb turns on, the fault must be in the top part, otherwise the bottom part. In either case, the next test will certainly reveal the fault, as there are only two options left. For test \textbf{D} the same principle applies, but with the top and bottom switched.
}

Note that the worked example is only included for the re-introduction of the domain after the learning phases. Before the learning phases, these paragraphs are omitted. The other domains are introduced in a similar way.

\begin{figure}[h]
    \centering
    \begin{subfigure}{0.95\linewidth}
        \centering
        \includegraphics[width=0.95\linewidth]{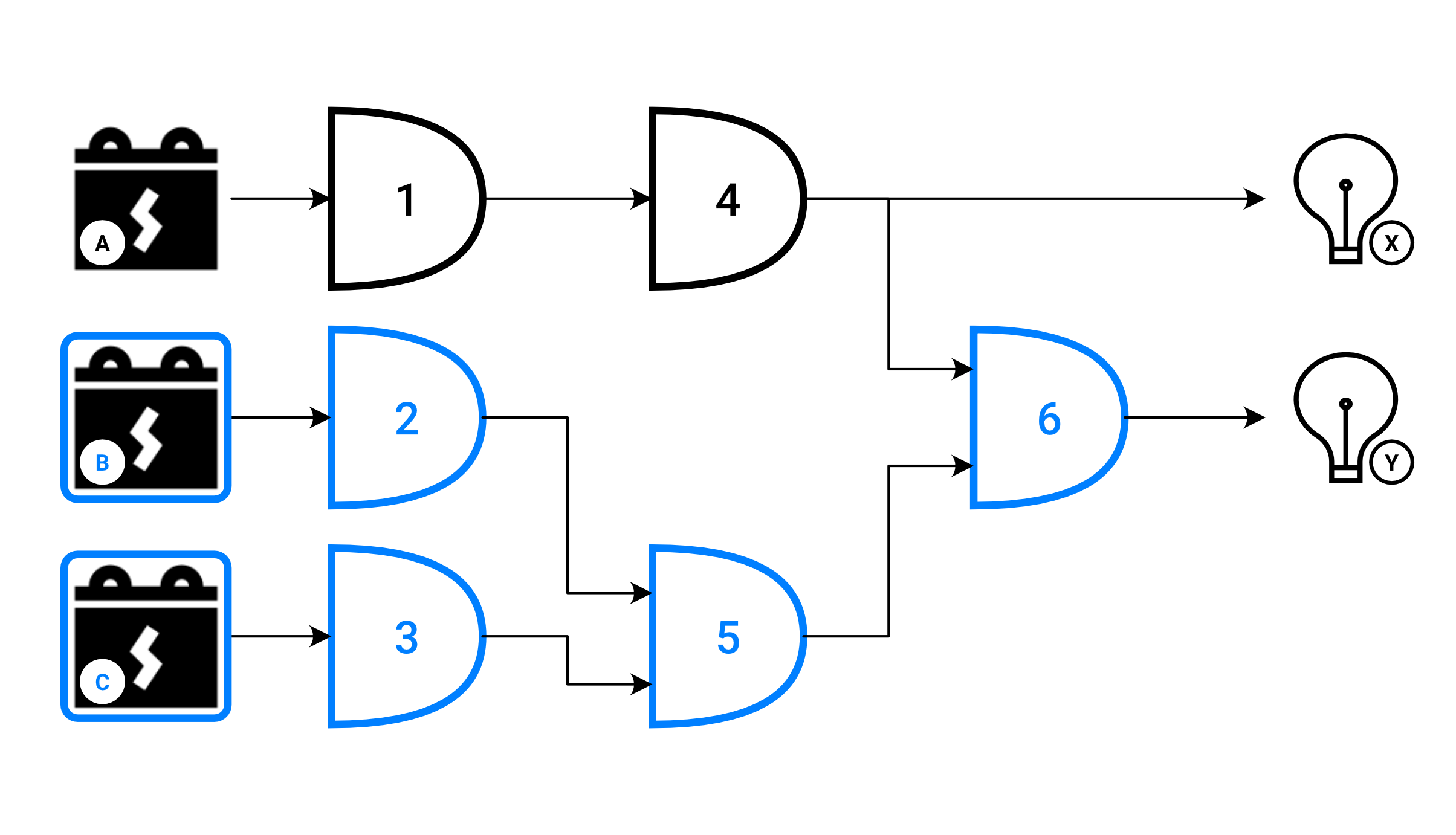}
        \caption{Visual feedback given to $H_1$.}
    \end{subfigure}
    \begin{subfigure}{0.95\linewidth}
        \centering
        \includegraphics[width=0.95\linewidth]{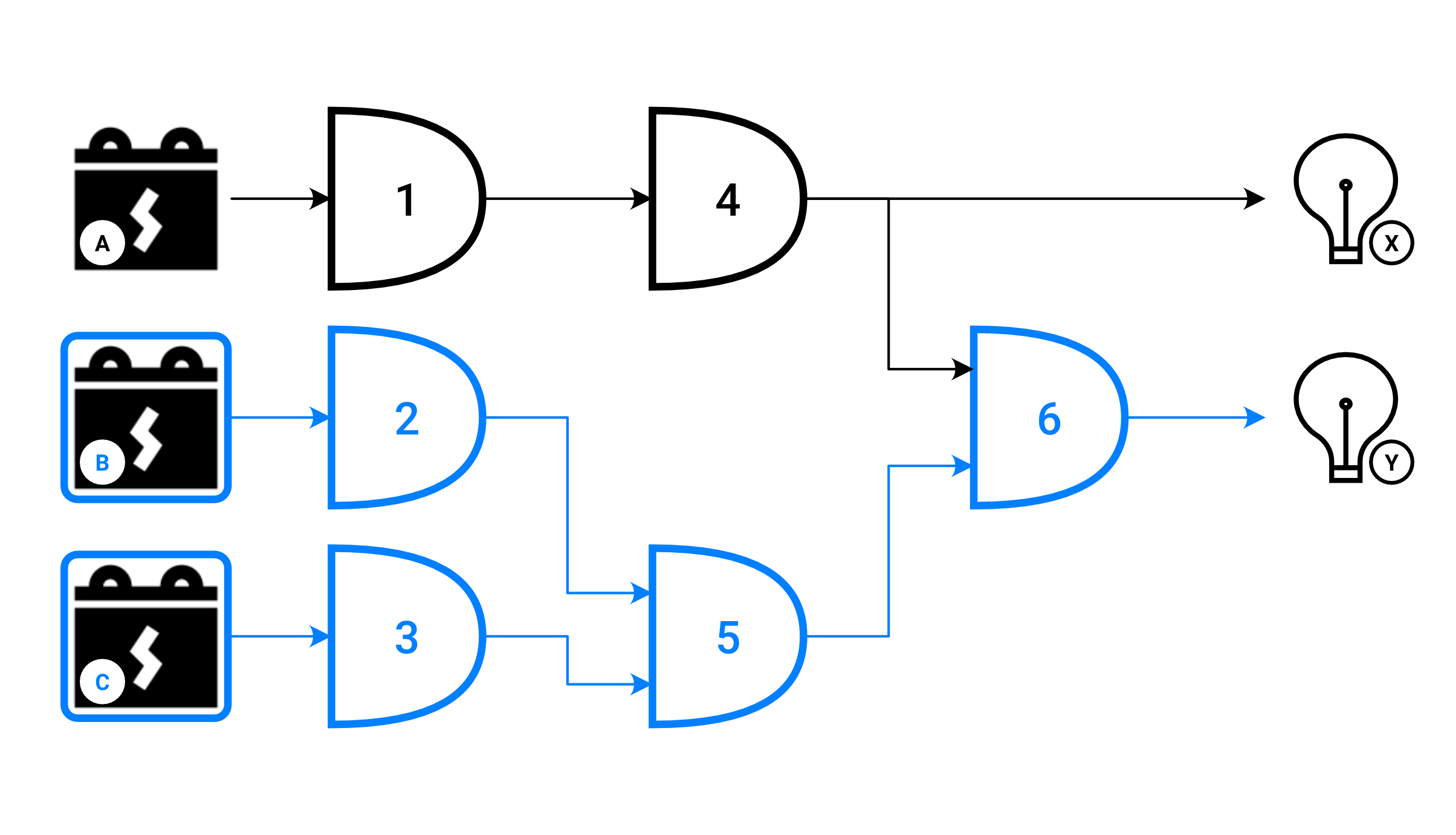}
        \caption{Visual feedback given to $H_2$}
    \end{subfigure}
    \caption{Participants see this circuit without any highlighted nodes or edges during learning phase 1.}
    \label{fig:learning_01}
\end{figure}

\paragraph{First Learning Phase.} The first learning phase aims to teach participants to identify dominators in a directed graph. Participants see the graph in Figure~\ref{fig:learning_01} without any highlights and the following task description:

\textit{Find all gates and batteries which are exclusively a source of power for lightbulb Y, that is they don’t supply current to any other lightbulb than Y.}

Participants are offered the numbered gates 1 through 6, the batteries A, B, and C as well as the lightbulbs X and Y as multiple choice options.

\begin{figure}[h]
    \centering
    \begin{subfigure}{0.95\linewidth}
        \centering
        \includegraphics[width=0.95\linewidth]{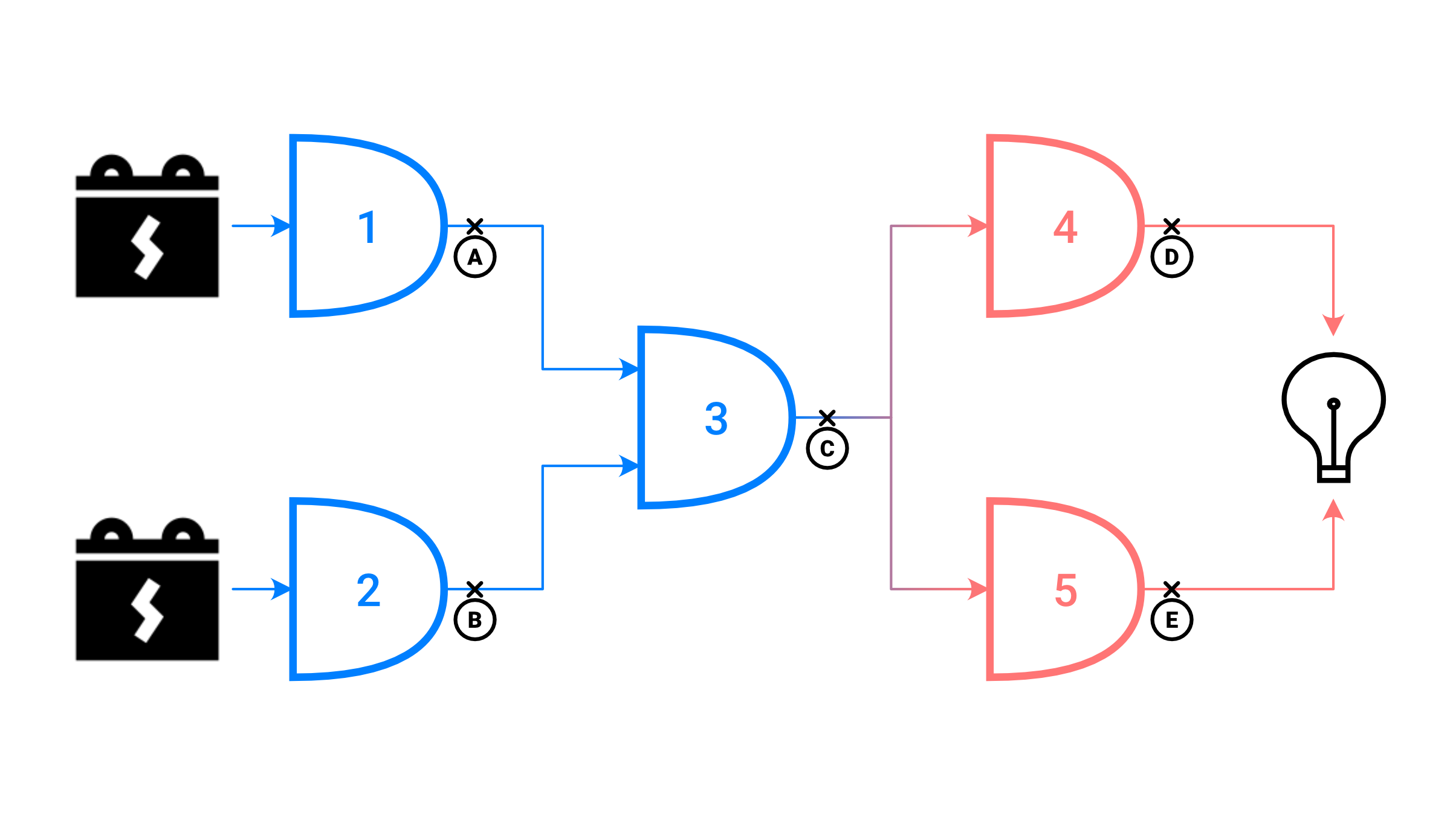}
        \caption{The first circuit shown to participants.}
    \end{subfigure}
    \begin{subfigure}{0.95\linewidth}
        \centering
        \includegraphics[width=0.95\linewidth]{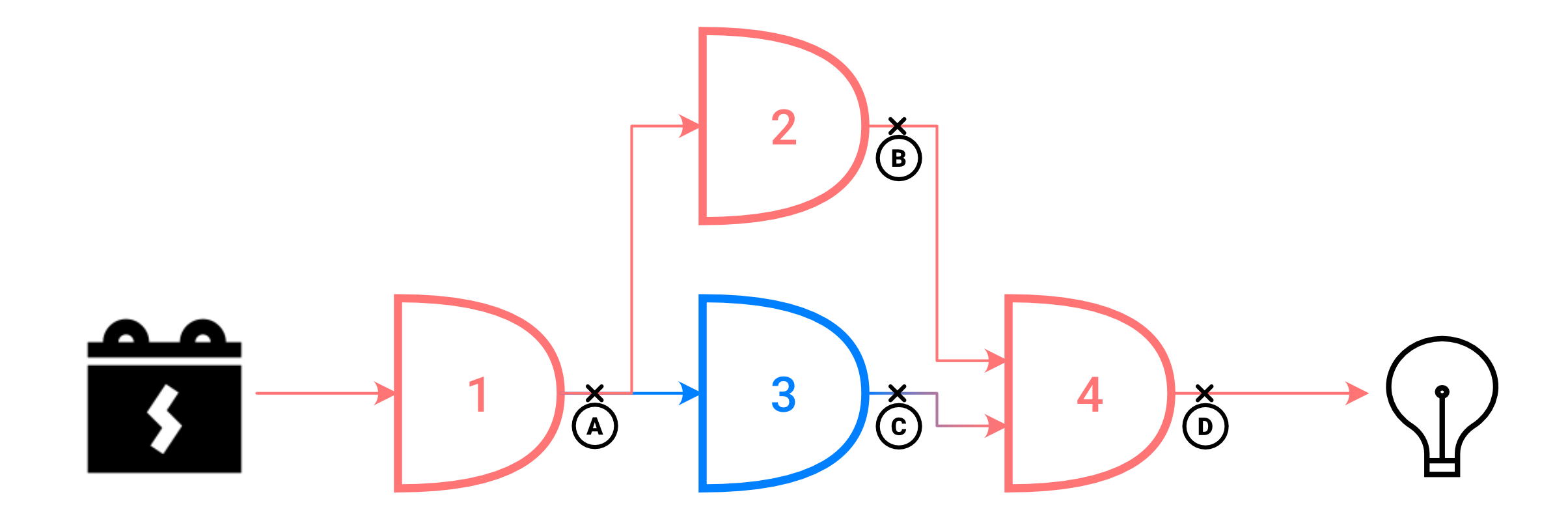}
        \caption{The second circuit shown to participants with the goal of illustrating bypasses.}
    \end{subfigure}
    \caption{Participants see these circuits without any highlighted nodes or edges during learning phase 2. These circuits contain the visual feedback for $H_2$. $H_1$ sees no highlighted edges just as it is demonstrated in Figure~\ref{fig:learning_01}.}
    \label{fig:learning_02}
\end{figure}

\paragraph{Second Learning Phase.} The second learning phase aims to teach participants to partition the graph. Participants see the graphs in Figure~\ref{fig:learning_02} without any highlights and the following task description:

\textit{
    The goal of performing a test is to divide the gates in the network into two groups: The ones which may be faulty, and the ones which we can be certain work fine. Remember: Test Points are cables labeled with capital letters. To perform a test, consider the following:
}

\textit{
    When you perform a test, you add a power supply to the selected cable. If the lightbulb turns on after supplying power, the fault must be somewhere \textbf{before} the test cable, as the network seems to work fine with power supplied at a point in the middle. If the lightbulb still doesn’t turn on after supplying power, the fault must be \textbf{after} the test point. Since we also know that there is only one fault, this means all previous nodes work fine.
}

We offer two different circuits to illustrate the edge-case of a bypass around a test point. For each numbered gate in the circuit, participants are given a single choice option between \textit{certainly working}, \textit{potentially faulty} and \textit{don't know}.

\begin{figure}[h]
    \centering
    \begin{subfigure}{0.95\linewidth}
        \centering
        \includegraphics[width=0.95\linewidth]{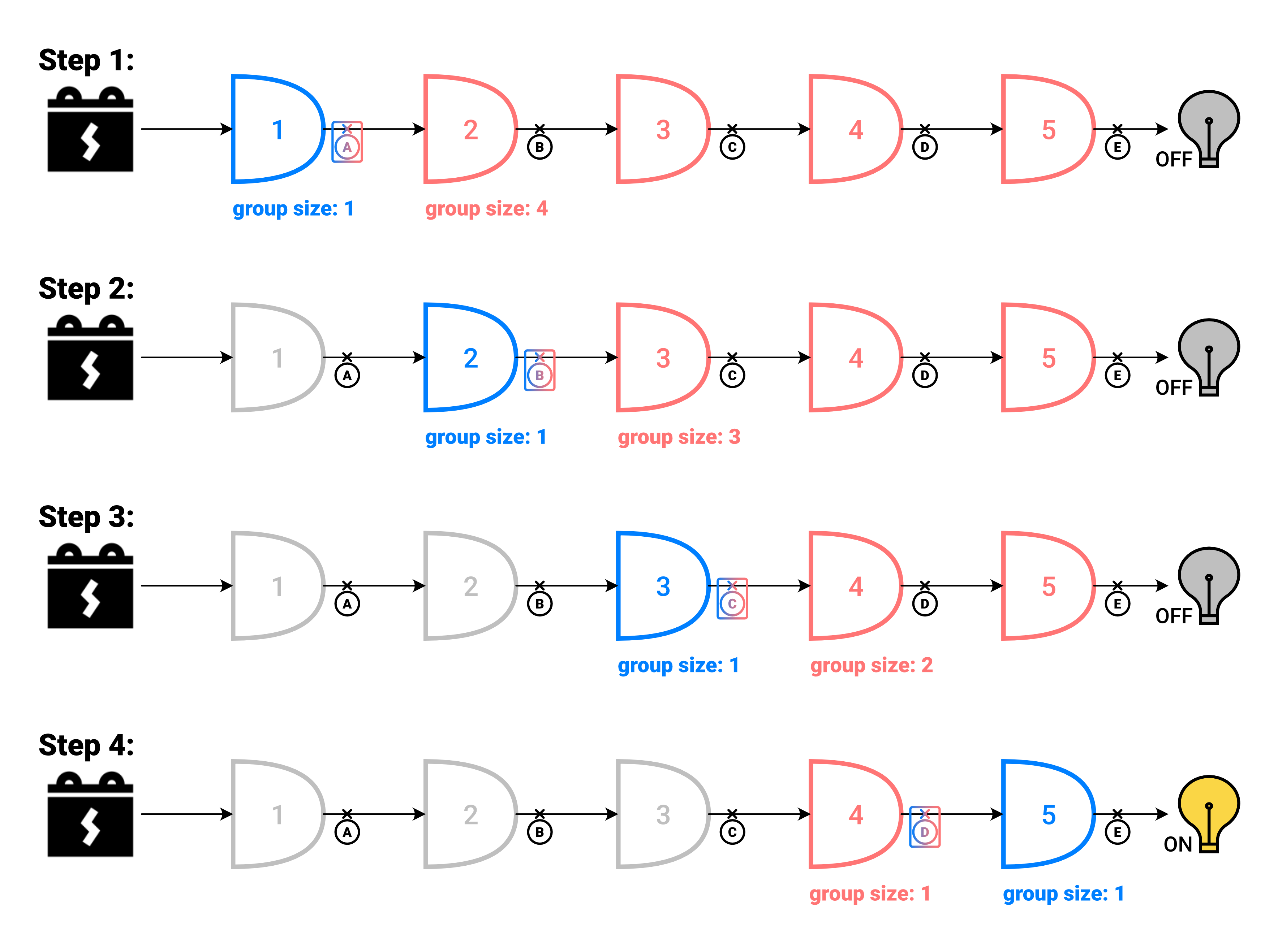}
        \caption{Option 1.}
    \end{subfigure}
    \begin{subfigure}{0.95\linewidth}
        \centering
        \includegraphics[width=0.95\linewidth]{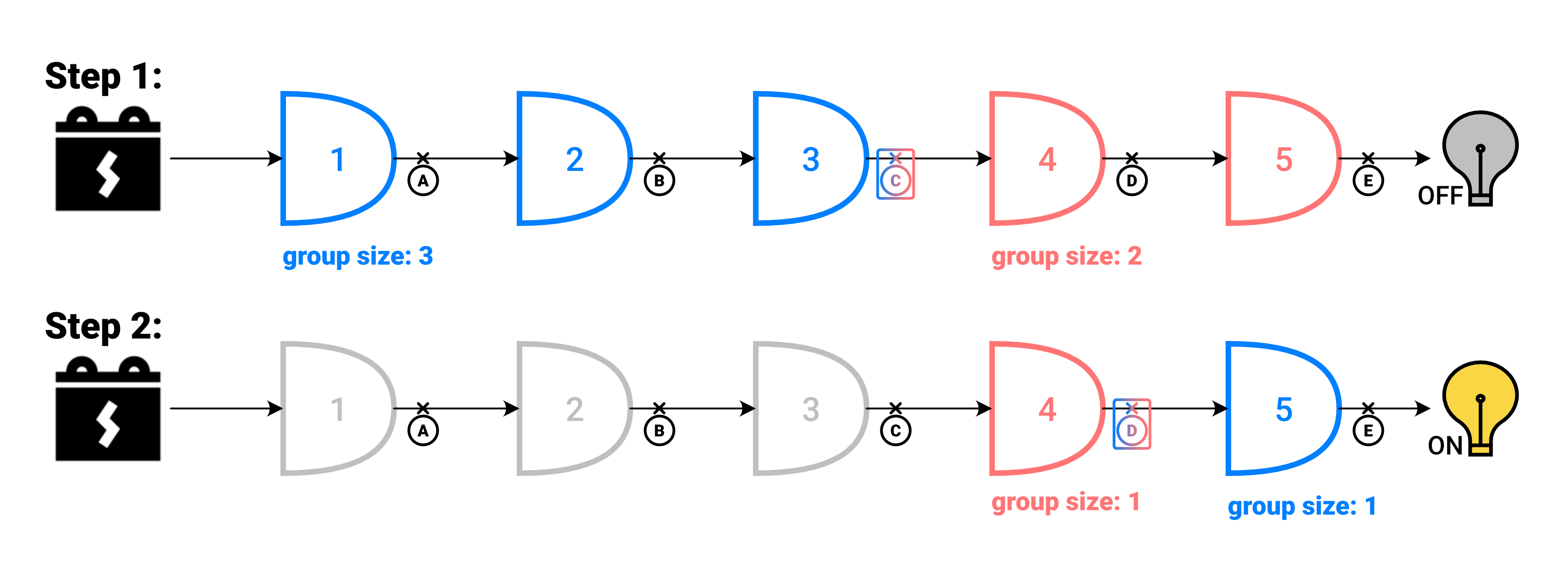}
        \caption{Option 2.}
    \end{subfigure}
    \caption{Two traces for solving a simple linear circuit as seen by $H_2$. $H_1$ does not see the group size annotations. Apart from that, the visual presentation is identical.}
    \label{fig:learning_03}
\end{figure}

\paragraph{Third Learning Phase.} The third learning phase aims to teach participants to select the locally optimal test based on partition sizes. Participants see the two solution traces shown in Figure~\ref{fig:learning_03} with the following task description:

\textit{
    Consider the following approaches to finding the fault in the same simple circuit. Familiarise yourself with the two strategies and think about which of the two is better suited for finding faults quickly and reliably.
}

Participants are then given a single choice between the two options displayed. Unlike the first two learning phases, participants do not receive the pipeline-generated explanations with the feedback, but directly after the two traces.

\begin{figure}[h]
    \centering
    \begin{subfigure}{0.95\linewidth}
        \centering
        \includegraphics[width=0.95\linewidth]{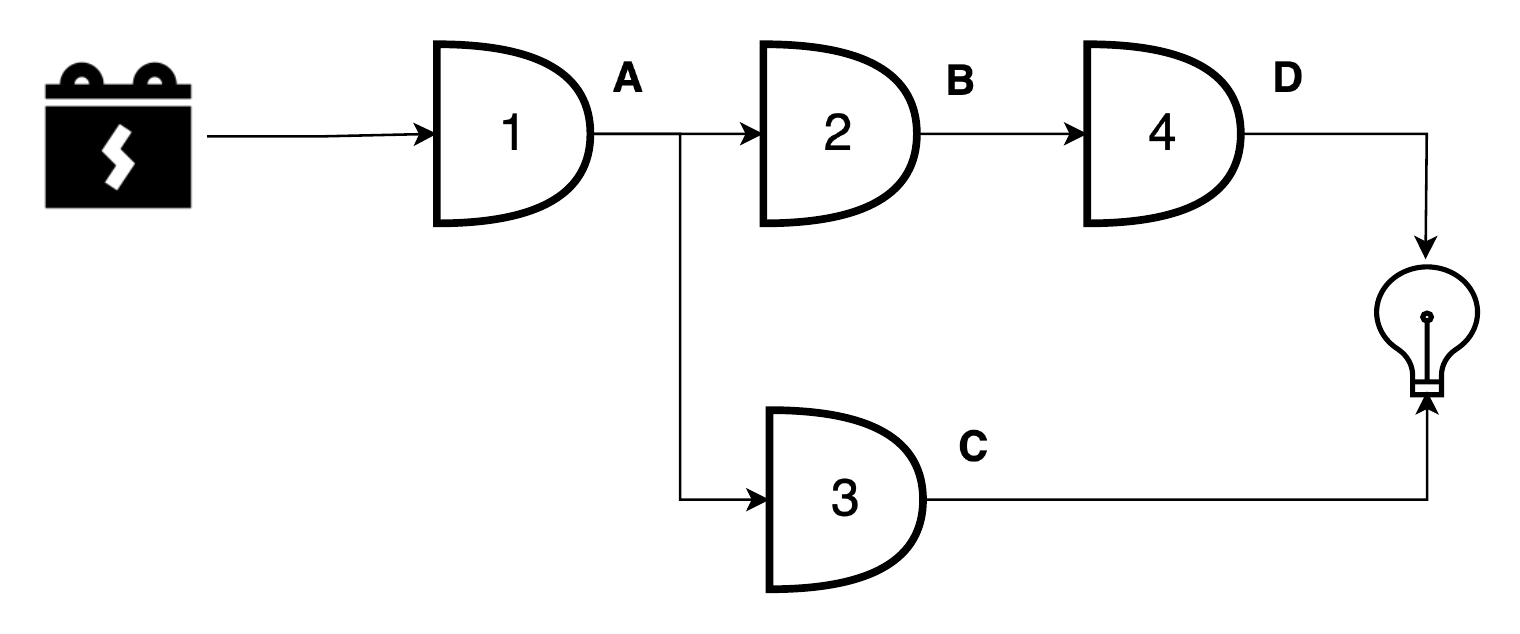}
        \caption{Circuit domain.}
    \end{subfigure}
    \begin{subfigure}{0.95\linewidth}
        \centering
        \includegraphics[width=0.95\linewidth]{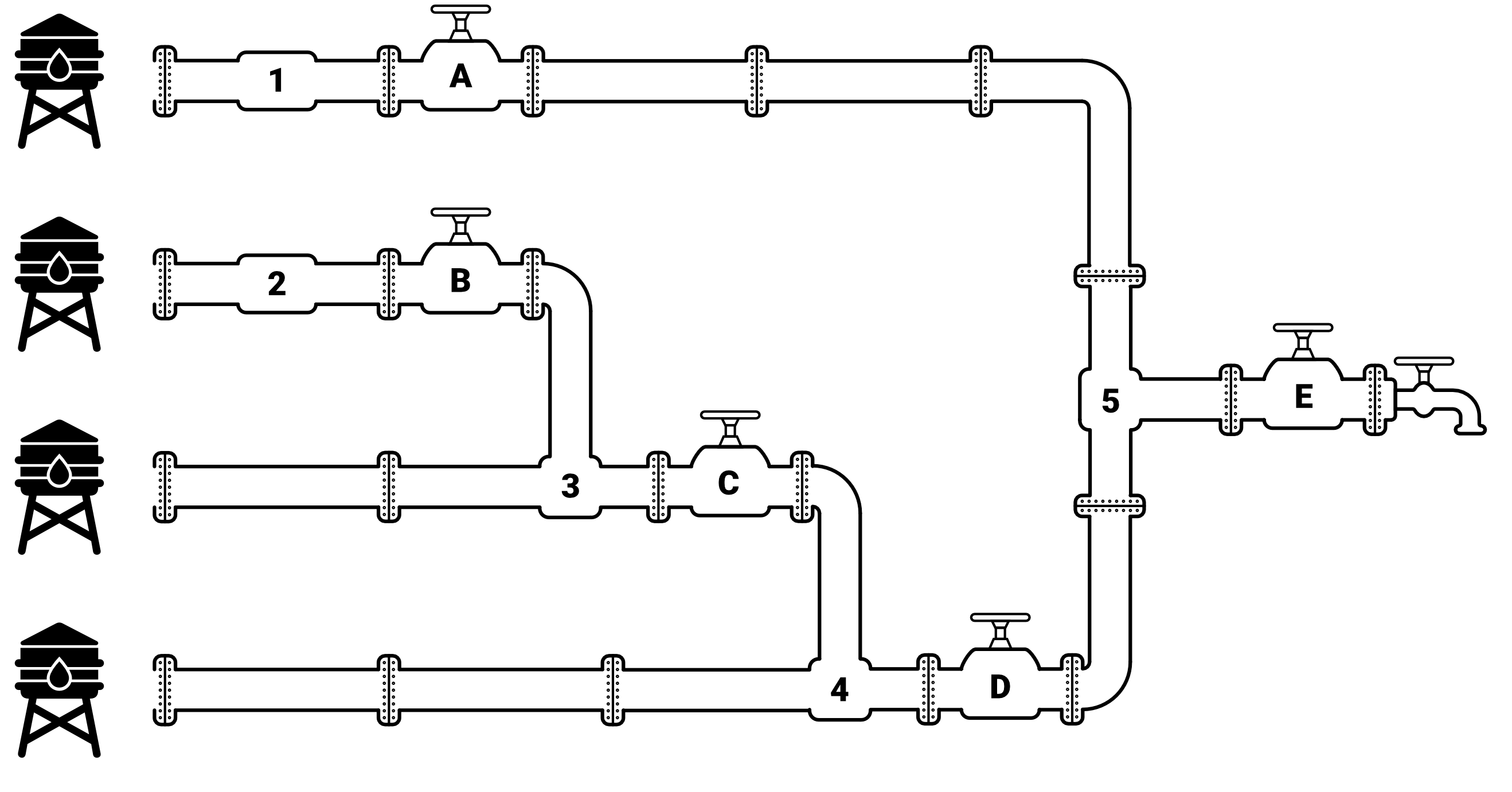}
        \caption{Water flow domain.}
    \end{subfigure}
    \begin{subfigure}{0.95\linewidth}
        \centering
        \includegraphics[width=0.95\linewidth]{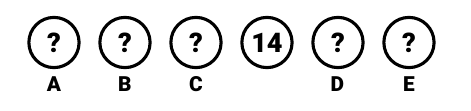}
        \caption{List binary search domain.}
    \end{subfigure}
    \caption{Exemplary graphs used during the test phase.}
    \label{fig:trials}
\end{figure}

\paragraph{Trial Items.} Figure~\ref{fig:trials} shows exemplary trial graphs shown to participants for each domain. 


\paragraph{Data Quality.} To ensure data quality, three comprehension questions assessed participants' basic understanding of the first domain. To detect potential AI assistance, we used attention check questions with hidden instructions that would only be revealed if participants attempted to copy the page content directly into an LLM. Participants who failed either comprehension or attention checks beyond a predetermined threshold were not allowed to continue the experiment.

\section{Selected Explanations and Case Studies}
\label{appendix:selected_explanations_and_case_studies}

\subsection{Selected Explanations}
\label{appendix:selected_explanations}

Here we include explanations selected from the pipeline for the empirical study. Each explanation is labelled by task, condition and reasoning model. In these conditions, the pipeline uses the same two coding models, Qwen2.5 coder 14B and StarCoder V2 15B.

\begin{cbox}[title={Explanation for Task 1 (pipeline + AP + GC, DeepSeek R1)},colframe={blue!80}]
To solve this task, we use two main strategies:\\

1. \textbf{Path Checker}: This checks if a gate can send power to a specific lightbulb's connection point. It works by:\\
   - Directly matching if the gate is the lightbulb's connection point itself\\
   - Following each outgoing arrow from the gate to see if any path leads to the lightbulb's connection point\\

2. \textbf{Exclusive Source Verifier}: This confirms a gate only powers one specific lightbulb (Y) by:\\
   - Using the Path Checker to verify it powers lightbulb Y\\
   - Checking it doesn't power any other lightbulb (using the Path Checker for all other lightbulbs)\\

The solution involves:\\
1. Finding all gates that power lightbulb Y using the Path Checker\\
2. For each candidate gate, using the Exclusive Source Verifier to confirm it doesn't power any other lightbulb\\
3. Keeping only gates that pass both checks\\

This ensures we find gates that exclusively send power to lightbulb Y and nowhere else.\\
\end{cbox}

\begin{cbox}[title={Explanation for Task 2 (pipeline + NP, DeepSeek R1)},colframe={blue!80}]
To solve the task of dividing gates into two groups based on a test at point C (one group that may be faulty and one that is certainly working), follow these steps:\\

\textbf{Relevant Strategies}\\
1. \textbf{Same circuit check}:  \\
   Confirm if a gate is in the same circuit as point C. This ensures we only consider gates affected by the test.\\

2. \textbf{Exclusively powers check}:  \\
   Determine if point C exclusively powers a gate. If true, the gate relies solely on C for power.\\

3. \textbf{Not exclusively powers check}:  \\
   Determine if point C does not exclusively power a gate. This means the gate gets power from other sources besides C.\\

4. \textbf{Partition operation}:  \\
   Split the list of gates into two groups using the above checks:\\
   - Group 1: Gates in the same circuit as C and exclusively powered by C.\\
   - Group 2: Gates in the same circuit as C but not exclusively powered by C.\\

\textbf{Step-by-Step Application}\\
1. \textbf{Identify relevant gates}:  \\
   Use the "same circuit check" to collect all gates in the same circuit as point C. Ignore gates outside this circuit.\\

2. \textbf{Apply the partition operation}:  \\
   - For Group 1 (certainly working if C passes the test):  \\
     Use the "exclusively powers check" to find gates powered only by C. If C works correctly, these gates cannot be faulty.\\
   - For Group 2 (may be faulty):  \\
     Use the "not exclusively powers check" to find gates powered by C and other sources. These gates could still be faulty even if C works.\\

3. \textbf{Result}:  \\
   The partition operation directly outputs the two groups:\\
   - Group 1: Gates that are certainly working (if C passes the test).\\
   - Group 2: Gates that may be faulty (regardless of C’s test outcome).\\

\textbf{Key Insight}\\
The test at point C helps isolate responsibility:  \\
- Gates exclusively powered by C depend entirely on C’s functionality.  \\
- Gates with multiple power sources might have issues beyond C.  \\
This division pinpoints where faults could exist based on the test results.\\
\end{cbox}

\begin{cbox}[title={Explanation for Task 3 (pipeline + AP + GC, Claude 3.7 Sonnet)},colframe={blue!80}]
Choosing the Most Appropriate Test to Find a Fault\\

To decide which test is most appropriate for finding a fault quickly and reliably in an electrical circuit, we need to use several strategies:\\

\textbf{Relevant Strategies}\\

1. \textbf{Test Point Evaluator}: Assesses different test points to determine which one provides the most useful information.\\

2. \textbf{Circuit Divider}: Splits the circuit into sections based on where test points are located.\\

3. \textbf{Group Size Calculator}: Counts how many gates fall into each group when a particular test is applied.\\

4. \textbf{Optimal Test Finder}: Identifies which test creates the most balanced division of gates into groups.\\

\textbf{How to Apply These Strategies}\\

1. First, identify all possible test points in the circuit (these are usually labeled with letters).\\

2. For each test point, use the Circuit Divider to determine how it would split the gates into two groups:\\
   - Gates that might be faulty if this test point is used\\
   - Gates that would be confirmed as working properly\\

3. Use the Group Size Calculator to count how many gates would be in each group for each potential test point.\\

4. Apply the Optimal Test Finder to compare all possible test points and find the one that:\\
   - Creates the most balanced division (where the smaller group is as large as possible)\\
   - Minimizes the number of gates you would need to check after the test\\
   - Provides the clearest information about where the fault might be\\

5. The test point that creates the most balanced division is usually the best choice because:\\
   - It eliminates the maximum number of possibilities with a single test\\
   - It divides the remaining possibilities into manageable groups\\
   - It helps you find the fault with the fewest additional tests\\

This approach is similar to a binary search strategy, where you try to eliminate half of the possibilities with each test. By choosing the test point that creates the most balanced division, you can find the fault in the circuit with the minimum number of tests, saving time and effort.
\end{cbox}

\subsection{Case Study LLM Responses}
\label{appendix:case_study}

Owing to the large volume of generated explanations, we present case studies highlighting typical errors identified in LLM responses. These errors highlight the importance of combining human evaluation and automated LLM evaluations. The most common error type involves proposing fundamentally different approaches from both the reference answer and the ILP-learned logic programs, which we attribute to LLM hallucination. Hallucination is a well-documented tendency for language models to generate plausible-sounding but factually incorrect content.

For example, the following response from direct prompting of DeepSeek R1 provides a specific algorithm for identifying the groups by starting from the test point and working forwards or backwards (\hl{highlighted in yellow}). This proposed algorithm incorrectly assumes that any test point serves as the primary dominator, implying that exactly half the circuit gates would eventually pass through it. However, this assumption does not hold in general circuit topologies, so the strategy is not technically correct.

\begin{cbox}[title={Explanation for Task 3 (RM + GC, DeepSeek R1)},colframe={blue!80}]
To solve this task of choosing the best test point to find the fault quickly and reliably, use these strategies:\\

Strategy 1: \textbf{Focus on the Lightbulb Path}\\
Identify the specific path from the battery to the lightbulb that isn't working. Only consider test points along this path, as they directly affect the faulty lightbulb. Ignore test points not on this path.\\

\hl{Strategy 2: \textbf{Count Gates Before and After}\\
For each test point on the path:  \\
- Count the gates between the battery and the test point (""gates before"").  \\
- Count the gates between the test point and the lightbulb (""gates after"").  }\\

Strategy 3: \textbf{Choose the Most Balanced Test}\\
For each test point, compare its ""gates before"" and ""gates after"" counts:  \\
- Take the \textbf{larger} of these two numbers for each test point.  \\
- Select the test point where this larger number is the \textbf{smallest} among all options.  \\

How to Apply:  \\
1. Use \textbf{Focus on the Lightbulb Path} to list all test points between the battery and the faulty lightbulb. \\ 
2. For each test point, apply Count Gates Before and After to get two numbers.  \\
3. Apply Choose the Most Balanced Test:  \\
   - Compare the larger number from each test point.  \\
   - Pick the test point with the smallest larger number.  \\

Why This Works:  \\
This test point splits the path into segments as evenly as possible. After testing:  \\
- If the lightbulb turns on, the fault is in the ""gates before"" segment.  \\
- If it stays off, the fault is in the ""gates after"" segment.  \\
By choosing the test point where the larger segment is as small as possible, you minimize the worst-case number of gates to check next. This saves time and ensures reliability.\\
\end{cbox}

Despite this fundamental error, the explanation received mixed ratings: 8 from DeepSeek R1, 6 from Claude 3.7 Sonnet, and 9 from o3-mini, illustrating how superficially coherent but incorrect explanations can still receive favourable scores from some judges. 

The second common error type involves overgeneralised responses that lack sufficient specificity. While the example below correctly references some key components from the ILP-learned logic programs, it fails to identify the central concept of how to construct proper partitions in detail. This type of error demonstrates how LLMs can produce responses that appear knowledgeable by incorporating some relevant terminology, yet remain unactionable due to missing the core algorithmic concepts.

\begin{cbox}[title={Explanation for Task 3 (pipeline + NP + GC, Claude 3.7 Sonnet)},colframe={blue!80}]
Choosing the Best Test Point to Find a Fault\\

To find a fault quickly and reliably, we need to choose the most appropriate test point in the circuit.\\

Relevant Tools\\

- \textbf{Circuit Group Divider}: Separates gates into groups based on specific criteria.\\

- \textbf{Group Size Calculator}: Determines how many gates are in each group.\\

- \textbf{Optimal Division Finder}: Identifies the best way to divide gates into groups to efficiently locate faults.\\

- \textbf{Best Test Selector}: Determines which test point will be most effective for finding faults.\\

Strategy to Solve This Task\\

1. \textbf{Evaluate All Possible Test Points}: We need to consider each available test point in the circuit.\\

\hl{2. \textbf{For Each Test Point}:\\
   - Use the Circuit Group Divider to separate gates into two groups:\\
     * Gates that might be faulty if the test is successful\\
     * Gates that must be working properly if the test is successful}\\

3. \textbf{Calculate Group Sizes}: For each potential test point, determine how many gates fall into each group using the Group Size Calculator.\\

\hl{4. \textbf{Find the Optimal Division}: The best test point will divide the gates in a balanced way, so that:\\
   - Neither group is too large\\
   - The difference between group sizes is minimized\\
   - This makes it easier to narrow down the location of the fault}\\

5. Select the Best Test: Use the Best Test Selector to choose the test point that creates the most balanced division.\\

Applying This Strategy\\

To apply this strategy:\\

1. Identify all available test points in the circuit.\\

2. For each test point, use the Circuit Group Divider to separate the gates into two groups.\\

3. Use the Group Size Calculator to determine how many gates are in each group for each test point.\\

4. Use the Optimal Division Finder to compare all the different divisions and identify which one is most balanced.\\

5. The Best Test Selector will then choose the test point that creates this optimal division.\\

6. Perform the test at this selected point first, as it will give you the most information about where the fault might be, allowing you to find the problem more quickly and with fewer tests.\\

This approach ensures you're using your testing resources efficiently to locate the fault with minimal effort.
\end{cbox}

\section{ILP and Reference Materials}
\label{appendix:ilp_and_reference}

\subsection{Hopper Input and ILP-Learned Episodes}
\label{appendix:hopper_input}

\paragraph{Episode Descriptions.}
In episode 1, the system invents \texttt{exclusively\_powers}, which determines whether the output from a given gate flows exclusively through another gate on its path to the lightbulb. This strategy is analogous to identifying dominators in a directed graph \cite{prosser_applications_1959}: a node $B$ dominates node $A$ with respect to target $T$ if every path from $A$ to $T$ passes through $B$.

In episode 2, the system uses \texttt{exclusively\_powers} to invent \texttt{partition}, which splits the circuit gates into two groups given a test point: those exclusively powered through it, and those that are not. Episode 3 builds on these to learn \texttt{partition\_sizes}, computing the size of each partition. In episode 4, \texttt{optimal\_partition\_sizes} compares partition sizes to find the most balanced split. Finally, episode 5 invents \texttt{optimal\_test}, identifying the test point achieving the most balanced partition.

Hopper receives Prolog programs representing example, background knowledge and language bias as input. As part of the background knowledge for Hopper, each circuit is represented as relations representing the circuit components and their connectivity. 

\begin{cbox}[title={Circuit Example}]
\begin{scriptsize}
\begin{verbatim}
gate(1). gate(2). gate(3). gate(4). gate(5).

test_point_label(1, output_a). 
test_point_label(2, output_b).
test_point_label(3, output_c). 
test_point_label(4, output_d).
test_point_label(5, output_e).

is_connected(0, 1). is_connected(0, 2).
is_connected(1, 3). is_connected(2, 3).
is_connected(3, 4). is_connected(3, 5).
is_connected(4, lightbulb).
is_connected(5, lightbulb).
\end{verbatim}
\end{scriptsize}
\end{cbox}

We declare the predicates that would appear in the program and define the types, as well as which program arguments are inputs and outputs. We refer to these as language bias and use language bias similar to that from \cite{cropper_learning_2021,ilp_experiments}. Our second-order background knowledge was built from \cite{purgal_learning_2022}.

\begin{cbox}[title={Background Knowledge}]
\begin{verbatim}
find_all(P, A, L):- 
    findall(H, call(P, A, H), L).

all(P, [H|T], C) :- 
    call(P, H, C), !,
    all(P, T, C).
all(_, [], _).

empty_list([]).

is_equal(A, A).
same_circuit(A, B) :- gate(A), gate(B), 
    N is A // 100, M is B // 100, N == M.
size(L, S) :- is_list(L), length(L, S).

not_list(A) :- not(is_list(A)).

pair(A,B,[A,B]) :- not_list(A), not_list(B).
larger_min_size(A, B, A) :-
    is_list(A),
    is_list(B),
    length(A, 2),
    length(B, 2),
    min_list(A,M1),
    min_list(B,M2),
    max(M1, M2, M1).
larger_min_size(A, B, B) :-
    is_list(A),
    is_list(B),
    length(A, 2),
    length(B, 2),
    min_list(A,M1),
    min_list(B,M2),
    max(M1, M2, M2).

min(A, B, C) :- min_list([A,B],C).
max(A, B, C) :- max_list([A,B],C).

fold(P,Acc,[H|T],Out) :- 
    call(P,Acc,H,Inter),!,
    fold(P,Inter,T,Out).
fold(_P,Acc,[],Acc).

map(P,[H|T],[H1|T1]) :- 
    call(P,H,H1),!,
    map(P,T,T1).
map(_P,[],[]).

empty_partition_sizes([0, 0]).
\end{verbatim}
\end{cbox}

\begin{cbox}[title={Language Bias: (exclusively\_powers)}]
\begin{verbatim}
enable_recursion.

% Predicate declarations
head_pred(exclusively_powers,2). 
body_pred(is_equal,2).
body_pred(not_empty_list,1).
body_pred(is_connected,2).
body_pred(find_all,3,ho).
body_pred(all,3,ho).

% Types
type(exclusively_powers,(element,element)). 
type(is_equal,(element,element)).
type(not_empty_list,(list,)).
type(is_connected,(element,element)).
type(find_all,((element,element),element,list)).
type(all,((element,element),list,element)).

% Input-output signatures
direction(exclusively_powers,(in,out)). 
direction(is_equal,(in,out)).
direction(not_empty_list,(in,)).
direction(is_connected,(in,out)).
direction(find_all,((in,out),in,out)).
direction(all,((in,out),in,out)).

% Constraint the occurrence of predicates
occurFO(is_equal, 1).
occurFO(not_empty_list, 1).
occurFO(is_connected, 1).
occurHO(find_all,1).
occurHO(all, 1).
:-
    body_pred(X,_),
    #count{C,X,V: body_literal(C,X,_,V)} > Z, 
    occurFO(X,Z).
:- 
    body_pred(X,_,ho), 
    #count{C,Y,V: body_literal(C,Y,_,V),
                X=@honameparse(Y)} >Z, 
    occurHO(X,Z).

% Turn off invented predicates in HO
:- invented_ho_used(_,_).
\end{verbatim}
\end{cbox}

\begin{cbox}[title={Language Bias: (partition)}]
\begin{verbatim}
% Predicate declarations
head_pred(partition,3). 
body_pred(exclusively_powers,2).
body_pred(not_exclusively_powers,2).
body_pred(same_circuit,2).
body_pred(find_all,3,ho).

% Types
type(partition,(element,list,list)). 
type(exclusively_powers,(element,element)).
type(not_exclusively_powers,(element,element)).
type(same_circuit,(element,element)).
type(find_all,((element,element),element,list)).

% Input-output signatures
direction(partition,(in,out,out)). 
direction(exclusively_powers,(in,out)).
direction(not_exclusively_powers,(in,out)).
direction(same_circuit,(in,out)).
direction(find_all,((in,out),in,out)).

% Must use the following predicates
:- not body_literal(_,same_circuit,_,(0,1)).
:- not body_literal(_,exclusively_powers,_,(1,0)).
:- not body_literal(_,not_exclusively_powers,
                    _,(1,0)).
\end{verbatim}
\end{cbox}

\begin{cbox}[title={Language Bias: (partition\_sizes)}]
\begin{verbatim}
% Predicate declarations
head_pred(partition_sizes,2). 
body_pred(partition,3).
body_pred(size,2).
body_pred(pair,3).

% Types
type(partition_sizes,(element,list)). 
type(partition,(element,list,list)).
type(size,(list,element)).
type(pair,(element,element,list)).

% Input-output signatures
direction(partition_sizes,(in,out)). 
direction(partition,(in,out,out)).
direction(size,(in,out)).
direction(pair,(in,in,out)).    
\end{verbatim}
\end{cbox}

\begin{cbox}[title={Language Bias: (optimal\_partition\_sizes)}]
\begin{verbatim}
% Predicate declarations
head_pred(optimal_partition_sizes,2). 
body_pred(partition_sizes,2). 
body_pred(empty_partition_sizes,1).
body_pred(larger_min_size,3).
body_pred(map,3,ho).
body_pred(fold,4,ho).

% Types
type(optimal_partition_sizes,(list,list)). 
type(partition_sizes,(element,list)). 
type(empty_partition_sizes,(list,)).
type(larger_min_size,(list,list,list)).
type(map,((element,list),list,list)).
type(fold,((list,list,list),list,list,list)).

% Input-output signatures
direction(optimal_partition_sizes,(in,out)). 
direction(partition_sizes,(in,out)). 
direction(empty_partition_sizes,(out,)).
direction(larger_min_size,(in,in,out)).
direction(map,((in,out),in,out)).
direction(fold,((in,in,out),in,in,out)).

% Turn off invented predicates in HO
:- invented_ho_used(_,_).
\end{verbatim}
\end{cbox}

\begin{cbox}[title={Language Bias (optimal\_test)}]
\begin{verbatim}
% Predicate declarations
head_pred(optimal_test,2). 
body_pred(optimal_partition_sizes,2). 
body_pred(partition_sizes,2). 
body_pred(gate,1).
body_pred(test_point_label,2).

% Types
type(optimal_test,(list,element)). 
type(optimal_partition_sizes,(list,list)). 
type(partition_sizes,(element,list)). 
type(gate,(element,)).
type(test_point_label,(element,element)).

% Input-output signatures
direction(optimal_test,(in,out)). 
direction(optimal_partition_sizes,(in,out)). 
direction(partition_sizes,(in,out)). 
direction(gate,(out,)).
direction(test_point_label,(in,out)).

% Turn off invented predicates in HO
:- invented_ho_used(_,_).
\end{verbatim}
\end{cbox}

\subsection{Human Reference Answers}
\label{appendix:human_reference}

LLM judges received the following manually composed reference answers when scoring explanations. For the electric circuit domain, each of the three below references corresponds to a task discussed in Section~\ref{sub:llm} and~\ref{exp:usml_teaching}.

\begin{cbox}[title={Reference 1 (Electric Circuit)},colframe={blue!50}]
Find each gate, for which every outgoing cable eventually leads to a point in the circuit (either a gate or a lightbulb). Start from the point in question. For each incoming cable, check if the origin gate has cables to anything other than that point. If not, mark that gate. Now, check for each marked gate whether each outgoing cable leads to either the first point or another marked gate. If so, mark that gate as well. Repeat this, until there are no new gates to add.\\
\end{cbox}

\begin{cbox}[title={Reference 2 (Electric Circuit)},colframe={blue!50}]
The circuit must be divided in two groups: The first group contains all gates from which each outgoing cable eventually leads to the test point. The second group contains all other gates. To find the first group, start from the test point. For each incoming cable, check if the origin gate has cables to anything other than the test point. If not, add that gate to the group. Now, check for each gate added to the group whether each outgoing cable leads to either the test point or another gate in the group. If so, add that gate to the group as well. Repeat this until there are no more gates to add to the group. Finally, add all gates not in group 1 to group 2.\\
\end{cbox}

\begin{cbox}[title={Reference 3 (Electric Circuit)},colframe={blue!50}]
Choose the test for which the sizes of the two groups are the most balanced. For each possible test, note the sizes of the two resulting groups. Finally, choose the test, for which the smaller group is largest.\\
\end{cbox}

The past USML work \cite{Ai2021} features the Island Game, a specially designed two-player game isomorphic to Noughts and Crosses. The Island Game contains three islands, where each island has three territories, and every territory has one or more resources. A player wins when he or she controls either all territories on one island or three instances of the same resource. The nine territories resemble the nine cells in the Noughts and Crosses board. The game is played similarly to Noughts and Crosses through players’ turns until one player wins or no player can ever win. 

\begin{cbox}[title={Reference 1 (Game Playing, Island Game)},colframe={blue!50}]
Identify the move from the board, that can immediately create a triplet of the same resource type. Generally, you should locate resource types from your territories which you have two of the three. Taking the third resource would lead to the WINNING condition.\\
\end{cbox}

\begin{cbox}[title={Reference 2 (Game Playing, Island Game)},colframe={blue!50}]
Find a move that creates two pairs of different resources and make sure your opponent does not have any pairs. The idea is that you can build two threats instead of one, where one pair can easily be blocked if your opponent is playing optimally. In this case, your opponent could only block one of the two pairs and cannot win immediately, so you taking the third resource would lead to the WINNING condition.\\
\end{cbox}

\begin{cbox}[title={Reference 3 (Game Playing, Island Game)},colframe={blue!50}]
Choose a move that gives you one pair of resources and advantage over the next two moves. Your opponent would block your attempt to create a triplet of resources. You can follow up by creating two pairs of different types in your next turn. Your opponent could only block one of the two pairs, so you can reach the WINNING condition by taking the third resource.\\
\end{cbox}

The past USML work \cite{sequential_teaching} explored using ILP-learned logic programs to teach humans computational algorithms. In the paper, the authors focused on helping humans rediscover the concept of merge sort by explaining how to merge. Merge sort is a recursive sorting algorithm based on a divide-and-conquer approach. However, a bottom-up variant of the merge sort algorithm was learned for sorting positive integers. An isomorphic problem was designed to represent integer sequences by the weights of fruits in piles. The task requires arranging fruits from two boxes into a correct sequence on a conveyor belt while operating a balance scale to compare the weights of two fruits. 

\begin{cbox}[title={Reference 1 (Algorithm Discovery, Merge Sort)},colframe={blue!50}]
You should compare the two items from the boxes to find the lighter item. Then, append the lighter item to the conveyor belt. Afterwards, you should check if all remaining items are from one of the two boxes. In that case, you can safely add the rest of the items to the conveyor belt.\\
\end{cbox}

\subsection{Hand-Crafted Template Explanations}
\label{appendix:templates}

We adopted the textual explanation templates from \cite{Ai2021,sequential_teaching}. The parts in angle brackets correspond to where the visualisation would be. The placeholders in curly brackets would be replaced by evaluating the learned logic programs.

\begin{cbox}[title={Explanation 1 (Game Playing, Island Game)},colframe={blue!80}]
$<$Positive Example Move \{move\}$>$\\
This is a right move. You select this territory and obtain 1 triplet (\{resource\}).\\

$<$Negative Example Other Moves$>$\\
This is a wrong move. Contrast: No triplet.\\
\end{cbox}

\begin{cbox}[title={Explanation 2 (Game Playing, Island Game)},colframe={blue!80}]
$<$Positive Example Move \{move\}$>$\\
You select this territory and obtain 2 pairs (\{resource1\}, \{resource2\}). \\

$<$Positive Example Opponent's Move$>$\\
Opponent has no pair.\\

$<$Negative Example Other Moves Case 1$>$\\
This is a wrong move. Contrast: Not enough pair(s). \\

$<$Negative Example Other Moves Case 2$>$\\
This is a wrong move. Contrast: opponent has 1 pair. \\
\end{cbox}

\begin{cbox}[title={Explanation 3 (Game Playing, Island Game)},colframe={blue!80}]
$<$Positive Example Move \{move\}$>$\\
You select this territory and obtain 1 pair (\{resource\}). \\

$<$Positive Example Opponent's Move$>$\\
Opponent conquers and prevents you from getting a triplet (\{resource\}).\\ 

$<$Positive Example Next Move \{move\}$>$\\
You obtain 2 pairs (\{resource1\}, \{resource2\}) and opponent has no pair.\\

$<$Negative Example Other Moves$>$\\
This is a wrong move. Contrast: Not enough pair(s). \\

$<$Negative Example Next Moves$>$\\
This is a wrong move. Contrast: Not enough pair(s).  \\
\end{cbox}

\begin{cbox}[title={Explanation 4 (Algorithm Discovery, Merge Sort)},colframe={blue!80}]
Remaining Items are \{item1\} and \{item2\}.\\

$<$Positive Example Move \{item1\} then \{item2\}$>$\\
Item \{item1\} is lighter than item \{item2\}; append item \{item1\}. Append remaning item(s): \{item2\}.\\

$<$Negative Example Other Moves$>$\\
Item \{item1\} is lighter than item \{item2\} SO item \{item1\} should be appended.\\
\end{cbox}

\subsection{Prompt Templates}
\label{appendix:prompt_templates}

We used the following prompt templates for the experiment in Section~\ref{exp:pipeline_evaluation}.

\paragraph{Prompt Templates for Coding Models.}

\begin{promptbox}[title={System}]
You are an expert at explaining Prolog code to people with no programming background. Your job is to write clear, simple, and friendly explanations that help a complete beginner understand what a given Prolog program does. Assume the reader has no technical background. \\

Follow these steps:\\
1. Examine the structure and components of the Prolog program. (Note: invented programs are prefixed with inv, and higher-order programs use the ho identifier)\\
2. Translate the logic into plain English sentences that describe what each part does.\\
3. Explain the overall purpose of the program and what it's designed to achieve.\\
4. Avoid jargon. \\
5. Keep sentences short and accessible.
\end{promptbox}

\begin{promptbox}[title={User}]
Below is a Prolog program. Explain it to someone with no programming background. Avoid jargon and use simple language.\\

```prolog\\
\{prolog\}
```
\end{promptbox}
\bigskip

\paragraph{Prompt Templates for Reasoning Models.}

\begin{promptbox}[title={System}]
You are a reasoning assistant tasked with summarising how to apply effective strategies to solve a sequence of related tasks, for a person with no programming background. You are now given a task domain introduction and a set of sample explanations that illustrate the general context for the tasks and how the Prolog predicates function. \\

Your goal:\\
Produce a clear and concise explanation of how the task can be solved using the available predicates. Your explanation must be understandable to someone without any knowledge of programming or logic. \\

Your output should:\\
- Create a new name for each relevant predicate in solving the task, for reusing later.\\
- Teach the person by summarising how relevant predicates help achieve the task, using the new names.\\
- Use plain, accessible language without technical detail or code-like terminology.\\

DO NOT:\\
- Use predicate names, Prolog syntax, or variables\\
- Use jargons, analogies, metaphors, or comparisons\\

Please read the following:\\

[Task Domain Introduction]\\
\{domain\_context\}\\

[The End of Task Domain Introduction]\\

[Sample Explanations]\\
\{samples\}\\

[The End of Sample Explanations]\\
\end{promptbox}

\begin{promptbox}[title={System (Direct Prompting)}]
You are a reasoning assistant tasked with summarising how to apply effective strategies to solve a sequence of related tasks, for a person with no technical background. You are now given a task domain introduction that illustrates the general context for the tasks. \\

Your goal:\\
Produce a clear and concise explanation of how the task can be solved using your strategies. Your explanation must be understandable to someone without any knowledge of technical background. \\

Your output should:\\
- Create a new name for each relevant strategy component in solving the task, for reusing later.\\
- Teach the person by summarising how relevant strategy components help achieve the task, using the new names.\\
- Use plain, accessible language without technical detail.\\

DO NOT:\\
- Use jargons, analogies, metaphors, or comparisons\\

Please read the following:\\

[Task Domain Introduction]\\
\{domain\_context\}

[The End of Task Domain Introduction]
\end{promptbox}

\begin{promptbox}[title={System (No Global Context)}]
You are a reasoning assistant tasked with summarising how to apply effective strategies to solve a sequence of related tasks, for a person with no programming background. You are now given a set of sample explanations that illustrate how the Prolog predicates function. \\

Your goal:\\
Produce a clear and concise explanation of how the task can be solved using the available predicates. Your explanation must be understandable to someone without any knowledge of programming or logic. \\

Your output should:\\
- Create a new name for each relevant predicate in solving the task, for reusing later.\\
- Teach the person by summarising how relevant predicates help achieve the task, using the new names.\\
- Use plain, accessible language without technical detail or code-like terminology.\\

DO NOT:\\
- Use predicate names, Prolog syntax, or variables\\
- Use jargons, analogies, metaphors, or comparisons\\

Please read the following:\\

[Sample Explanations]\\
\{samples\}\\

[The End of Sample Explanations]
\end{promptbox}

\begin{promptbox}[title={User}]
You are given a task description and an example that illustrates the context for solving it. Identify the relevant predicates and summarise the strategies they represent. Explain how these strategies can be applied in solving the task below. Use the same names for the same predicates as established earlier. Avoid jargon and use simple language.\\

[Task Description]\\
\{description\}\\

[The End of Task Description]\\

[Example Type]\\
\{example\_type\}\\

[Example]\\
\{example\}\\

[The End of Example]\\
\end{promptbox}

\begin{promptbox}[title={User (Direct Prompting)}]
You are given a task description. Identify and summarise relevant strategies. Explain how these strategies can be applied in solving the task below. Use the same names for the same strategy components as established earlier. Avoid jargon and use simple language.\\

[Task Description]\\
\{description\}\\

[The End of Task Description]
\end{promptbox}

\begin{promptbox}[title={User (Direct Prompting + Local Context)}]
You are given a task description and an example that illustrates the context for solving it. Identify and summarise relevant strategies. Explain how these strategies can be applied in solving the task below. Use the same names for the same strategy components as established earlier. Avoid jargon and use simple language.\\

[Task Description]\\
\{description\}\\

[The End of Task Description]\\

[Example Type]\\
\{example\_type\}\\

[Example]\\
\{example\}\\

[The End of Example]
\end{promptbox}

\begin{promptbox}[title={User (No Local Context)}]
You are given a task description. Identify the relevant predicates and summarise the strategies they represent. Explain how these strategies can be applied in solving the task below. Use the same names for the same predicates as established earlier. Avoid jargon and use simple language.\\

[Task Description]\\
\{description\}\\

[The End of Task Description]
\end{promptbox}
\bigskip

\paragraph{Prompt Templates for Judging Models.}
\begin{promptbox}[title={System}]
Please act as an impartial judge and evaluate the quality of the responses provided by an AI assistant to the user question displayed below. A reference human answer and the assistant's answer will be provided to you. You will review the assistant's response based on the instructions provided to you. Your evaluation should consider factors such as the helpfulness, relevance, accuracy, depth, creativity, and level of detail of the response. Begin your evaluation by providing a short explanation. Be as objective as possible. After providing your explanation, please rate the response on a scale of 1 to 10 by strictly following this format: "[[rating]]", for example: "Rating: [[5]]".\\

[Instructions for Assistant’s Answer]\\
\{instructions\}\\

[The End of Instructions for Assistant’s Answer]
\end{promptbox}
\begin{promptbox}[title={User}]
[Question]\\
\{question\}\\

[The Start of Reference Answer]\\
\{answer\_ref\}

[The End of Reference Answer]\\

[The Start of Assistant’s Answer] \\
\{answer\} \\

[The End of Assistant’s Answer]
\end{promptbox}

\end{document}